\documentclass[10pt,twocolumn]{article}
\usepackage{hmi-preprint}
\usepackage{amsmath}
\usepackage{url}
\usepackage{booktabs}
\usepackage{nicefrac}
\usepackage{microtype}
\usepackage{marvosym}
\usepackage{multirow}
\usepackage{array}
\usepackage{anyfontsize}
\usepackage{placeins}
\usepackage{adjustbox}
\definecolor{cvprblue}{rgb}{0.21,0.49,0.74}
\hypersetup{colorlinks=true,linkcolor=cvprblue,citecolor=cvprblue,urlcolor=cvprblue}

\usepackage[font=small,labelfont=bf,labelsep=period,skip=6pt]{caption}

\usepackage{balance}

\DeclareRobustCommand{\modelname}{DiffWAM}

\newcommand{\nothing}[1]{}

\preprintlogos{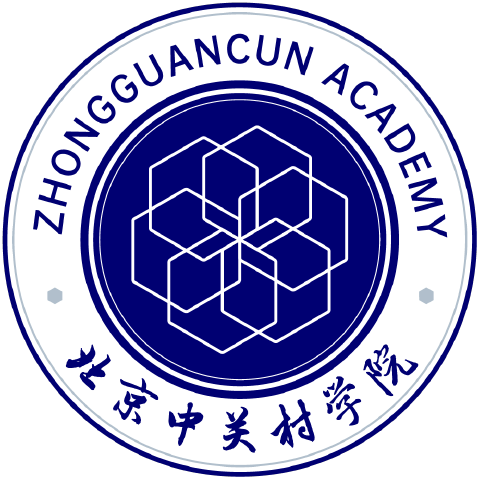}{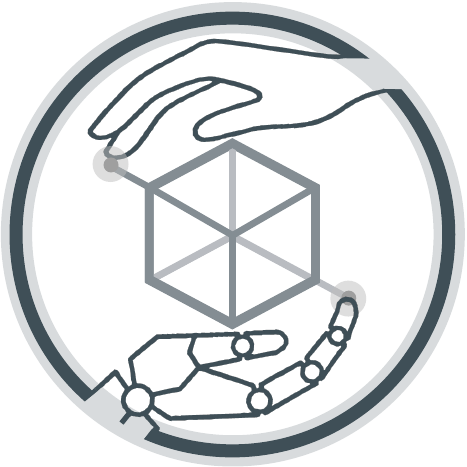}
\preprintdate{November 2026}
\runningtitle{\modelname{}: Learning World Action Model from Consecutive Visual Difference}
\preprintfooter{\textsuperscript{*}Equal contribution. \textsuperscript{\Letter} Corresponding author.}
\preprintlinkset
  {https://zgca-hmi-lab.github.io/DiffWAM}
  {https://github.com/ZGCA-HMI-Lab/DiffWAM}
  {}

\hypersetup{bookmarksdepth=2,pdftitle={DiffWAM: Learning World Action Model from Consecutive Visual Difference},pdfauthor={Xinheng Zhang, Haoyu Xiong, Chao Jing, Xiaohan Zhang, Linshan Li, Hongbang Fan, Xingyu Chen}}

\AtBeginDocument{\raggedbottom}
\newcommand{\resulttablefont}{\fontsize{8.5}{10}\selectfont}

\title{\modelname{}: Learning World Action Model from Consecutive Visual Difference}
\author[1,2,*]{Xinheng Zhang}
\author[5,3*]{Haoyu Xiong}
\author[6,3]{Chao Jing}
\author[1]{Xiaohan Zhang}
\author[1]{Linshan Li}
\author[5,3]{Hongbang Fan}
\author[3,4\Letter]{Xingyu Chen}
\affil[1]{College of Information Science and Electronic Engineering, Zhejiang University}
\affil[2]{Polytechnic Institute, Zhejiang University\qquad\protect\textsuperscript{3}Zhongguancun Academy}
\affil[4]{Zhongguancun Institute of Artificial Intelligence\qquad\protect\textsuperscript{5}Institute of Automation, Chinese Academy of Sciences}
\affil[6]{School of Intelligent Robotics and Advanced Manufacturing Innovation, Fudan University}

\makeatletter
\@namedef{@sep5}{\protect\\[3pt]}
\makeatother

\preprintteaser[fig:head]{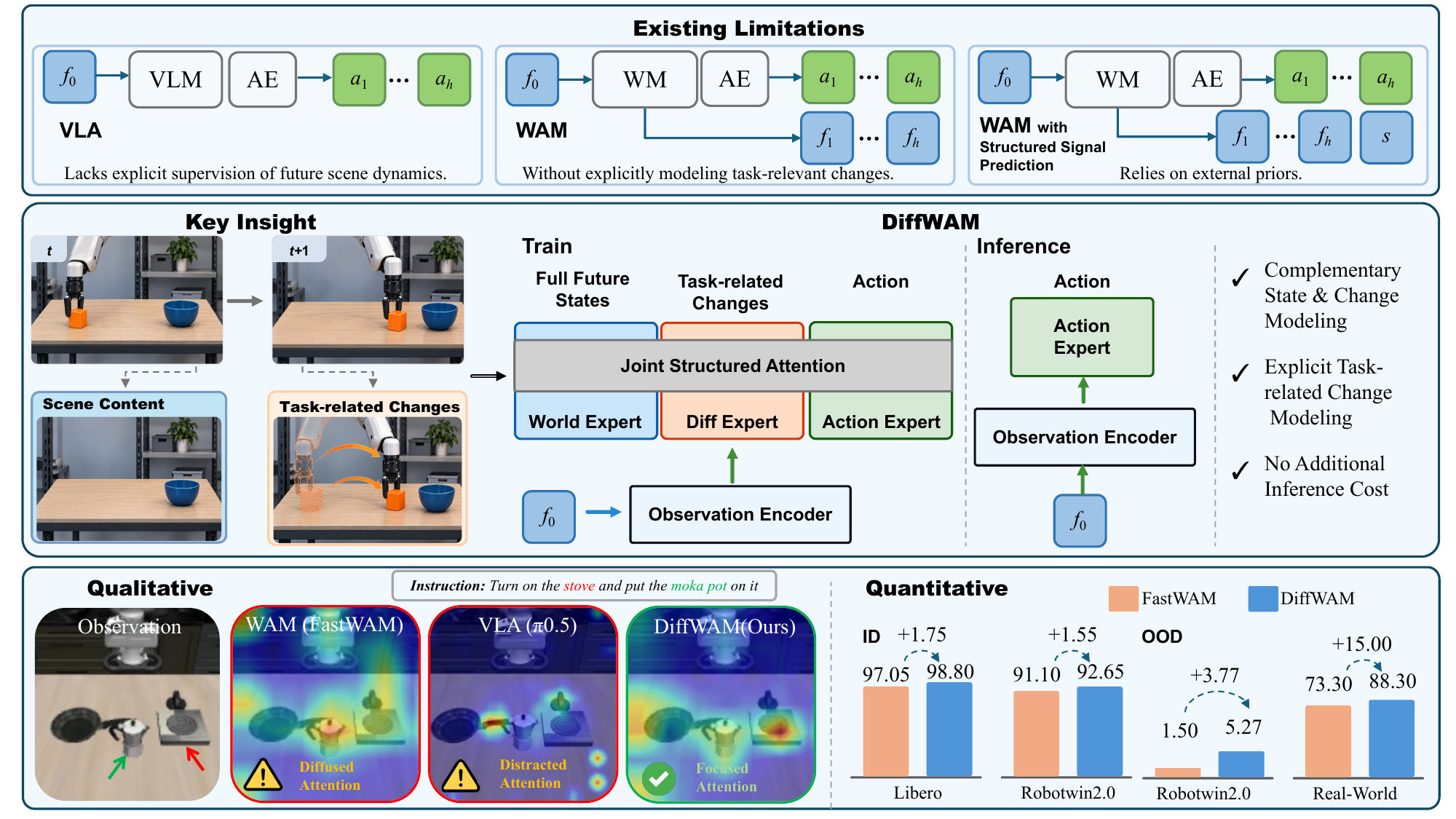}{\textbf{Overview of \modelname{}.} \textbf{Top: Existing limitations.} The illustrated VLA paradigm lacks explicit supervision of future scene dynamics; WAM predicts future states without explicitly distinguishing task-related changes; and WAM with Structured Signal Prediction (WAM-s) relies on external priors that may be unreliable. \textbf{Middle: Key insight and method.} Consecutive observations contain persistent scene content and task-related changes, motivating complementary future-state and visual-difference prediction. \modelname{} jointly trains World, Diff, and Action Experts under structured attention. Only the Observation Encoder and Action Expert are retained at inference, with no additional inference cost. \textbf{Bottom: Qualitative and quantitative results.} Attention visualizations suggest more focused attention on task-relevant regions than the illustrated VLA and WAM baselines. Success-rate comparisons with FastWAM demonstrate improvements on in-distribution LIBERO and RoboTwin 2.0 benchmarks, as well as out-of-distribution RoboTwin 2.0 and real-world evaluations.}

\begin{document}
\onecolumn
\maketitle
\twocolumn
\begin{abstract}
World Action Models (WAMs) have emerged as an important paradigm in robot learning, jointly predicting future scene states and robot actions from current observations and instructions. However, future states entangle task-related changes with large amounts of task-irrelevant content, such as static backgrounds, which weakens the model's ability to capture changes relevant to task execution. To address this, we propose Difference World Action Model (\modelname{}), a Mixture-of-Experts (MoE) framework that enhances world representations by jointly learning future-state prediction and consecutive visual difference prediction. We define consecutive visual differences as differences between the latent representations of consecutive observations, providing an explicit learning signal for temporal variation. The framework comprises a World Expert for future frame prediction, an Action Expert for action generation, and a Diff Expert for predicting these latent differences along the trajectory. A structured unidirectional attention mask conditions the Diff, World, and Action Experts solely on the current-frame representation while blocking attention between experts. This design allows both the World and Diff Experts to be removed during inference, introducing no additional inference overhead. \modelname{} achieves state-of-the-art performance on the LIBERO and RoboTwin 2.0 benchmarks and substantially improves over FastWAM on real-world manipulation tasks. Specifically, it attains average success rates of 98.8\%, 92.65\%, and 88.3\%, outperforming the state-of-the-art baseline (i.e., FastWAM) by 1.75, 1.55, and 15.0 percentage points, respectively.
\end{abstract}

\section{Introduction}

Learning task-related representations is crucial for robotic manipulation, as a robot must understand scene content and changes relevant to task execution. Two main paradigms have emerged to this end. Vision Language Action models (VLAs)~\cite{bib:vlasurvey1,bib:vlasurvey2} map observations and instructions directly to actions, which is simple and efficient but lacks explicit modeling of scene evolution during task execution~\cite{bib:rt1,bib:rt2,bib:openvla,bib:openvla-oft,bib:pi0,bib:pi05,bib:groot-n1}. In contrast, World Action Models (WAMs)~\cite{bib:wamsurvey1,bib:wamsurvey2}, building on the predictive world modeling paradigm~\cite{bib:dreamerv3}, learn by predicting future scene states alongside robot actions, thereby acquiring a predictive understanding of task-related scene dynamics~\cite{bib:motus,bib:lingbot-va,bib:fastwam,flowwam}.

Despite recent progress in WAMs~\cite{bib:uwm,bib:dreamzero,bib:oawam,bib:lawam,bib:dswam}, a key question remains: \emph{is future-state prediction alone sufficient to learn world representations that capture task-related changes?} As illustrated in Figure~\ref{fig:head}, much of the scene content persists across consecutive observations, while task execution involves changes such as robot motion and object interactions. The full-state prediction objective supervises these changes together with persistent content, without providing a separate learning signal for temporal variation. Thus, the representation of task-related changes relies on what the model learns implicitly from predicting complete future states.

This observation suggests a principle of visual foresight: a WAM should model not only what the future scene will look like, but also how it changes over time. Future-state prediction captures scene structure, object appearance, and spatial relationships, while consecutive visual difference prediction provides a separate target for temporal variation. We define a \emph{consecutive visual difference} as the difference between the latent representations of two consecutive observations. Predicting this quantity, also referred to as \emph{diff prediction}, provides explicit supervision for temporal variation without requiring additional motion annotations. By jointly modeling future states and their consecutive visual differences, a WAM can develop a more comprehensive world representation that integrates scene understanding with task-related dynamics for action generation.

In this paper, we propose \modelname{}, a Mixture-of-Experts (MoE) framework that jointly learns future-state prediction, visual difference prediction, and action generation. A shared Observation Encoder encodes the current visual observation into world representations, which are used by a World Expert to predict future states, a Diff Expert to predict differences between consecutive frame latents, and an Action Expert to generate actions. A structured unidirectional attention mask allows each expert to attend to the current-observation representations while blocking attention between experts. This architecture integrates state and visual difference modeling into the world representations used for action generation, while keeping the prediction pathways separate. At inference, both the World and Diff Experts are removed, retaining only the Observation Encoder and Action Expert without additional inference overhead. Across LIBERO~\cite{bib:libero}, RoboTwin 2.0~\cite{bib:robotwin2}, and real-world manipulation tasks, \modelname{} achieves average success rates of 98.8\%, 92.65\%, and 88.3\%, respectively, outperforming FastWAM~\cite{bib:fastwam} by 1.75, 1.55, and 15.0 percentage points.

Our contributions are summarized as follows:
\begin{itemize}
    \item We introduce joint future-state and visual difference modeling in WAMs, enabling world representations to capture both scene content and task-related dynamics for action generation.
    \item We propose \modelname{}, a Mixture-of-Experts framework with a shared Observation Encoder and dedicated World, Diff, and Action Experts. A structured unidirectional attention mask separates the expert pathways, allowing the World and Diff Experts to be removed at inference without additional overhead.
    \item We validate \modelname{} on LIBERO, RoboTwin 2.0, and real-world manipulation tasks, demonstrating its effectiveness across simulated and physical settings. \modelname{} reduces the optimization steps required to reach 95\% average success on LIBERO by 28.1\%. Ablation studies validate the benefits of joint state and visual difference modeling.
\end{itemize}

\section{Related Work}

Existing embodied policy models can be broadly grouped into three paradigms according to their model outputs and supervision: VLA, WAM with Future Prediction, and WAM with Structured Signal Prediction. We detail these three paradigms below.

\subsection{VLA}

Vision-Language-Action models directly generate actions from visual observations and language instructions, learning representations for action generation without an explicit future-state prediction objective. This paradigm formulates the problem as learning a conditional action policy:
\begin{equation}
    \pi_{\theta}\!\left(a_{1:h} \mid o_0, l\right),
    \label{eq:action-only-policy}
\end{equation}
where $o_0$ denotes the current observation, $l$ the language instruction, and $a_{1:h}$ the action sequence over a horizon of $h$ steps. Such policies typically couple a VLM backbone with an action expert, directly decoding actions from observation and instruction~\cite{bib:vlasurvey3,bib:vlasurvey4}.

Representative works following this paradigm include RT-2~\cite{bib:rt2}, OpenVLA~\cite{bib:openvla}, $\pi0$~\cite{bib:pi0}, Octo~\cite{bib:octo}, Diffusion Policy~\cite{bib:diffusion-policy}, RDT-1B~\cite{bib:rdt-1b}, and Xiaomi-Robotics-0~\cite{bib:xiaomi-robotics-0}. These methods directly map observations to actions, and VLM-based approaches can benefit from pretrained semantic representations. However, learning directly from demonstrations may also capture spurious correlations, potentially contributing to compounding errors and reduced generalization~\cite{bib:vlasurvey5}.

\subsection{WAM with Future Prediction}

WAMs with future prediction jointly model future visual states and actions, extending action generation with an explicit representation of scene evolution. Conditioned on the current observation and language instruction, they model the joint distribution:
\begin{equation}
    p_{\theta}\!\left(o_{1:h}, a_{1:h} \mid o_0, l\right),
    \label{eq:future-prediction-policy}
\end{equation}
where $o_{1:h}$ denotes the predicted future frames and $a_{1:h}$ the corresponding action sequence. Rather than modeling environmental dynamics implicitly, these methods explicitly predict future frames to ground action generation~\cite{bib:wamsurvey3,bib:wamsurvey4}.

Representative works following this paradigm include Motus~\cite{bib:motus}, LingBot-VA~\cite{bib:lingbot-va}, Fast-WAM~\cite{bib:fastwam}, GigaWorld-Policy~\cite{bib:gigaworld-policy}, IRASim~\cite{bib:irasim}, VideoVLA~\cite{bib:videovla}, Video Prediction Policy~\cite{bib:video-prediction-policy}, and WorldVLA~\cite{bib:worldvla}. Future visual prediction connects scene evolution with action generation~\cite{bib:dust,bib:flare}. Nevertheless, accurately predicting temporally coherent future frames in high-dimensional pixel or latent space remains challenging. For methods that condition action decoding directly on predicted future states, prediction errors may propagate into action generation~\cite{bib:dreamvla}. Moreover, predicting complete future states does not explicitly distinguish persistent scene content from inter-frame visual differences, leaving temporal variation to be learned implicitly within the state prediction objective. This can make it difficult to emphasize task-related changes over visually salient but task-irrelevant details~\cite{bib:cot-vla}.

\begin{figure*}[t]
    \centering
    \includegraphics[width=0.95\textwidth]{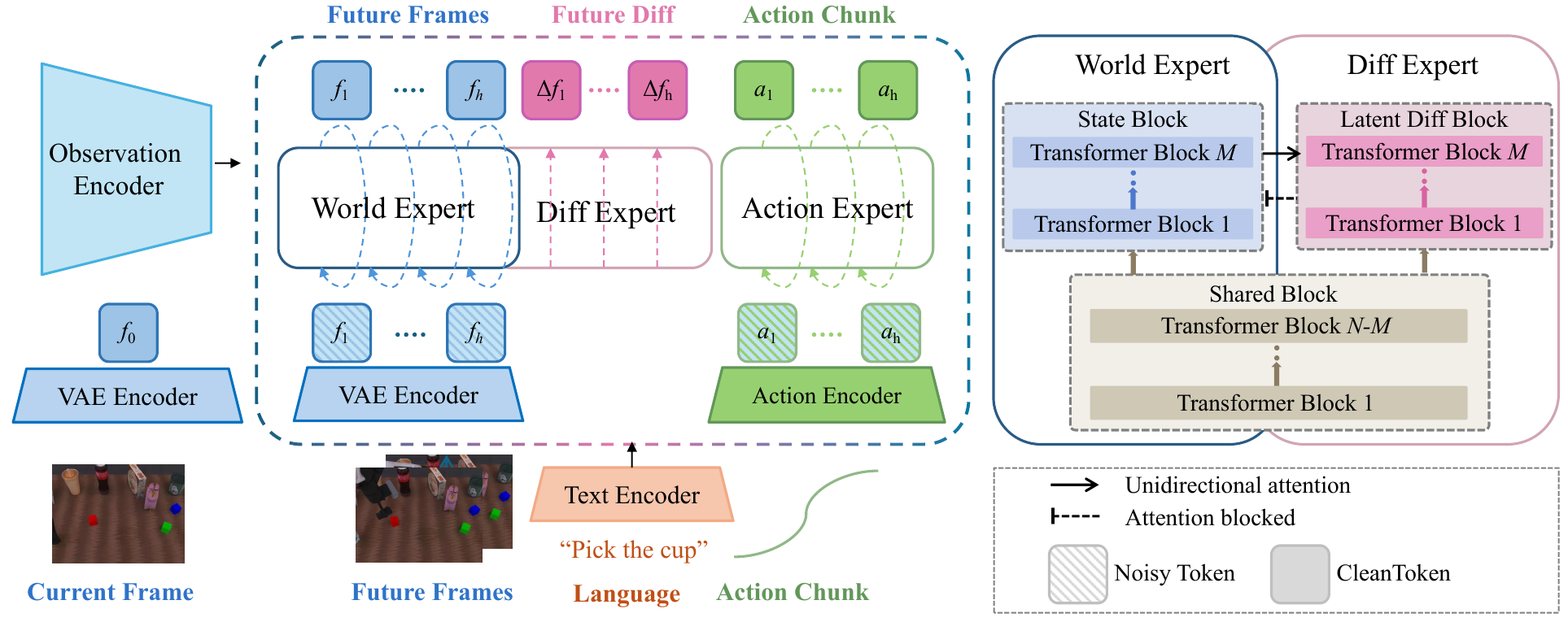}
    \caption{\textbf{Framework of \modelname{}.} The Observation Encoder provides current-observation representations to the World, Diff, and Action Experts. The World and Diff Experts receive the same noised future-frame latents, share their first $N-M$ layers, and use independent final $M$ layers for state generation through denoising and latent-difference prediction through regression, respectively. Structured unidirectional attention separates the expert pathways. At inference, both visual prediction experts are removed, retaining only the Observation Encoder and Action Expert without additional inference overhead.}
    \label{fig:principle}
\end{figure*}

\begin{figure}[t]
    \centering
    \includegraphics[width=\columnwidth]{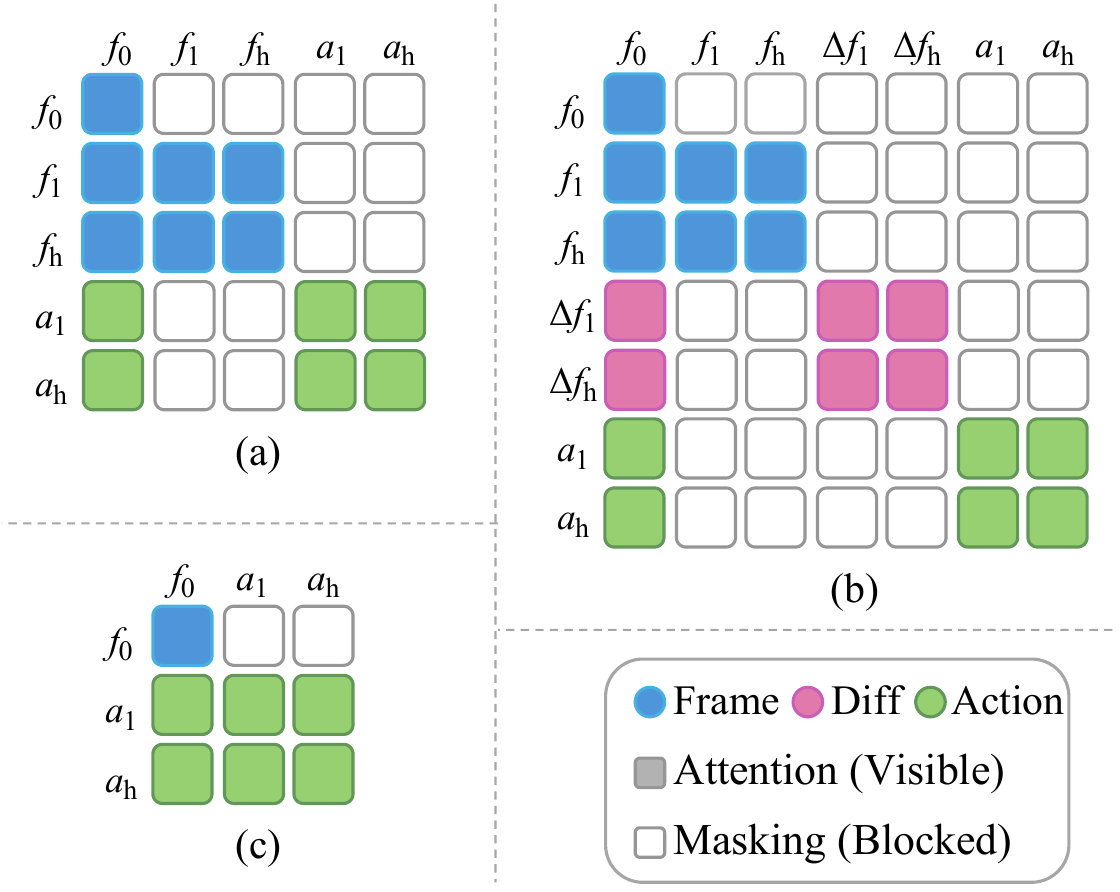}
    \caption{Structured attention masks in \modelname{}. (a) Attention in the first $N-M$ layers, where the World and Diff Experts share visual processing layers. (b) Attention in the last $M$ layers, with independent state and diff prediction pathways. (c) Observation and action pathways retained at inference. Each pathway attends to observation representations and its own tokens, while cross-attention between expert pathways is blocked.}
    \label{fig:mask}
\end{figure}

\subsection{WAM with Structured Signal Prediction}

Beyond future-state prediction, some WAMs incorporate structured prediction targets, such as depth, semantic segmentation, optical flow, or pretrained visual features, to model additional aspects of the scene. Formally, such methods introduce a structured representation $s$ and model
\begin{equation}
    p_{\theta}\!\left(o_{1:h}, s, a_{1:h} \mid o_0, l\right),
    \label{eq:structured-signal-prediction}
\end{equation}
where $o_{1:h}$ denotes the predicted future frames, $a_{1:h}$ the corresponding actions, and $s$ an auxiliary structured signal, such as a depth map, semantic segmentation, optical flow~\cite{flowwam,bib:flowdreamer}, or a pretrained feature field (e.g., DINOv3 features~\cite{bib:dinov3}). These targets capture geometric, semantic, or motion-related information, depending on the chosen signal. By supplementing future-state prediction with such structure, these methods introduce additional inductive biases into visual modeling.

Representative works include X-WAM~\cite{bib:x-wam}, AIM~\cite{bib:aim}, AGRA~\cite{bib:agra}, DeFI~\cite{bib:defi}, and MoLA~\cite{bib:mola}. These approaches specify particular geometric, semantic, or motion-related quantities as prediction targets. Their modeling emphasis therefore depends on the chosen signal, and targets obtained from external estimators may contain estimation errors. In contrast, \modelname{} directly models differences between consecutive frame latents alongside future states, integrating state and visual difference prediction within a WAM.

These paradigms provide different ways to connect visual modeling with action generation. Our work extends future-predictive WAMs by explicitly modeling both future states and inter-frame latent differences, aiming to capture scene content and task-related dynamics within a shared world representation.

\section{Method}
\subsection{Problem Formulation}

As illustrated in Figure~\ref{fig:principle}, \modelname{} jointly models future states, inter-frame visual differences, and actions conditioned on the current observation $o_0$ and a language instruction $l$. Let $f_i$ denote the VAE latent representation of observation $o_i$, $T$ the number of future visual states, and $H$ the action chunk length. The joint modeling objective is expressed in latent space as:
\begin{equation}
    p_{\theta}\!\left(f_{1:T}, \Delta f_{1:T}, a_{1:H} \mid f_0, l\right),
    \label{eq:dynamicwam-joint-distribution}
\end{equation}
where $f_{1:T}$ denotes the future visual latents, $a_{1:H}$ the action chunk, and $\Delta f_{1:T}$ the consecutive visual differences represented in latent space (labeled \emph{Future Diff} in Figure~\ref{fig:principle}), defined as
\begin{equation}
    \Delta f_i = f_i - f_{i-1}, \qquad i=1,\ldots,T.
    \label{eq:action-relevant-feature-change}
\end{equation}

We use \emph{visual difference prediction} for this auxiliary task and \emph{diff prediction} as its shorthand; the Diff Expert predicts the latent differences $\Delta f_i$. Future-state prediction captures scene content, while visual difference prediction explicitly supervises inter-frame variation. Jointly learning these complementary prediction tasks shapes the current-observation representation used for action generation. The joint objective does not require the Action Expert to consume predicted future states or latent differences; each expert is conditioned on the current-observation representation and language instruction. Latent differences describe visual variation and may also include nuisance changes, rather than exclusively encoding changes caused by robot actions.

\subsection{Model Architecture}

\textbf{Overview.} \modelname{} jointly learns future-state prediction, visual difference prediction, and action generation to enhance world representations. During training, the World and Diff Experts model future states and latent differences, respectively, while the Action Expert learns to generate actions from the current-observation representation. During inference, both the World and Diff Experts are removed, retaining only the Observation Encoder and Action Expert without additional inference overhead.

\textbf{Architecture.} As illustrated in Figure~\ref{fig:principle}, \modelname{} is a Mixture-of-Experts framework with a shared Observation Encoder and dedicated World, Diff, and Action Experts. The Observation Encoder and each expert contain $N$ Transformer layers. The World and Diff Experts share parameters in their first $N-M$ layers and use independent parameters in their last $M$ layers. The Observation Encoder maps the current visual latents into world representations that condition all three experts. The World Expert predicts future states, the Diff Expert predicts inter-frame latent differences, and the Action Expert generates the action chunk. A structured attention mask separates the expert pathways while allowing each expert to access the current-observation representation. Parameter sharing between the World and Diff Experts does not introduce cross-attention between their independent prediction pathways.

\textbf{Input Encoding.} \modelname{} inherits the pretrained text encoder and video VAE from Wan2.2-5B~\cite{bib:wan}. The task instruction $l$ is encoded with the T5 text encoder~\cite{bib:t5} and provides language conditioning through cross-attention. Images from multiple camera views are concatenated before being mapped into visual latents by the pretrained VAE~\cite{bib:vae}. The clean current-frame latents $f_0$ are processed by the Observation Encoder to provide the shared conditioning representation. During training, both the World and Diff Experts receive the same noised future-frame latents as input. The World Expert learns future-state generation through denoising, while the Diff Expert directly regresses latent differences, with targets computed from consecutive ground-truth visual latents. Actions $a_{1:H}$ are separately noised and embedded for the Action Expert.
    
\textbf{World Expert.} The World Expert models future states conditioned on the current-observation representation supplied by the Observation Encoder and the language instruction. Noised future-frame latents are processed by the first $N-M$ layers shared with the Diff Expert, followed by $M$ independent state-prediction layers. The World Expert learns the velocity field used for future-state generation through flow matching. Its prediction pathway is used during training and is removed at inference; the Observation Encoder is retained to provide the representation for action generation.

\textbf{Diff Expert.} The Diff Expert processes the same noised future-frame latents through the shared first $N-M$ layers, followed by an independent $M$-layer Latent Diff Block (LDB). Conditioned on the current-observation representation and language instruction, it directly regresses inter-frame latent differences rather than generating future states through denoising. Its independent pathway does not attend to the World or Action Expert's tokens, and its outputs are not passed to these experts. The regression targets are differences between consecutive ground-truth frame latents, including the transition from the current observation to the first future state. Visual difference modeling influences the shared representation during training, while the Diff Expert is removed at inference.

\textbf{Action Expert.} The Action Expert generates the action chunk through flow matching, conditioned on the current-observation representation supplied by the Observation Encoder and the language instruction. It does not directly receive predicted future states or diff outputs. State and visual difference modeling improve the representation used by the action pathway through joint training. At inference, the Action Expert operates with the Observation Encoder after both visual prediction experts have been removed.

\textbf{Structured Attention Mask.} As illustrated in Figure~\ref{fig:mask}, a structured unidirectional attention mask controls information flow between the current-observation pathway and the experts. Current-observation tokens do not attend to future-state, diff, or action tokens. Each expert may attend to the current-observation representation and tokens within its own pathway, while attention between different expert pathways is blocked. These restrictions concern interactions between visual and action token groups; language conditioning is supplied separately through cross-attention.

\noindent\textbf{(a)} In the first $N-M$ layers, the visual and action pathways follow a first-frame-conditioned attention scheme~\cite{bib:fastwam}: current-frame tokens attend to themselves, noised future-frame tokens attend to the current-frame representation and their own tokens, and action tokens attend to the current-frame representation and their own tokens. Future-frame and action tokens do not attend to each other. The World and Diff Experts share the visual processing layers at this depth.

\noindent\textbf{(b)} In the last $M$ layers, the World and Diff Experts use independent prediction pathways. Diff tokens attend to the current-frame representation and their own tokens, while the state and action pathways retain the attention restrictions in (a). Blocking cross-attention between these pathways prevents the Diff Expert from reading the World or Action Expert's token representations and prevents its outputs from entering their prediction pathways. This restriction does not exclude the noised future-frame input shared by the World and Diff Experts.

\noindent\textbf{(c)} At inference, only the current-observation and action pathways are retained. The Action Expert generates $a_{1:H}$ from the current-observation representation and language instruction without invoking future-state or diff prediction, introducing no additional overhead from the visual prediction experts.

\subsection{Joint Training Objective}

\modelname{} is trained with three losses: video flow matching~\cite{bib:flowmatching1} for future-state prediction, squared-error latent regression for difference prediction, and action flow matching~\cite{bib:flowmatching2} for action generation. Both flow branches are conditioned on the current-observation representation derived from $f_0$ and the language instruction $l$~\cite{bib:attena+}. In the following equations, conditioning on $f_0$ denotes access to this representation through the Observation Encoder.

\textbf{World Flow Matching Loss.} For each visual prediction step $i\in\{1,\ldots,T\}$, let $f_i$ denote the target future-frame latent and $\epsilon_i \sim \mathcal{N}(0,\mathbf{I})$ the sampled noise prior. We construct the interpolated sample along the probability path as $f_{i,\tau}=(1-\tau)\epsilon_i+\tau f_i$ with $\tau\in[0,1]$. Conditioned on the current-observation representation and the language instruction $l$, the World Expert is supervised to predict the corresponding velocity field:
\begin{equation}
    \begin{aligned}
        &\mathcal{L}_{\mathrm{world}} = \mathrm{E}_{i,\tau,\epsilon_i,f_i} \\
        &\quad \Bigl[\left\|v_{\theta}(f_{i,\tau},\tau\mid f_0,l)-(f_i-\epsilon_i)\right\|_2^2\Bigr],
    \end{aligned}
    \label{eq:world-flow-matching-loss}
\end{equation}
where $v_{\theta}$ is the predicted velocity field parameterized by $\theta$.

\textbf{Diff Regression Loss.} At each visual prediction step $i\in\{1,\ldots,T\}$, the Diff Expert receives the same noised future-frame latent $f_{i,\tau}$ as the World Expert and directly produces a prediction $\widehat{\Delta f}_i$ of the latent difference between consecutive observations. We use squared-error supervision in latent space, as also used for latent prediction in DINO-WM~\cite{bib:dinowm}. Here, the regression target $\Delta f_i=f_i-f_{i-1}$ is obtained by differencing consecutive ground-truth frame latents:
\begin{equation}
    \begin{aligned}
        &\mathcal{L}_{\mathrm{diff}} = \mathrm{E}_{i,\tau,\epsilon_i,f_{i-1},f_i} \\
        &\quad \Bigl[\left\|\widehat{\Delta f}_i-(f_i-f_{i-1})\right\|_2^2\Bigr],
    \end{aligned}
    \label{eq:delta-regression-loss}
\end{equation}
where $\widehat{\Delta f}_i$ is the LDB output at visual prediction step $i$, and $f_i-f_{i-1}$ is the target latent difference. The case $i=1$ uses the difference between the first future-frame latent and the current-frame latent.

\textbf{Action Flow Matching Loss.} Analogously, for each action step $i\in\{1,\ldots,H\}$, let $a_i$ be the ground-truth action and $\epsilon'_i\sim\mathcal{N}(0,\mathbf{I})$. We construct the interpolated sample as $a_{i,\tau}=(1-\tau)\epsilon'_i+\tau a_i$. Conditioned on the current-observation representation and the language instruction $l$, the Action Expert is trained to recover the target velocity field:
\begin{equation}
    \begin{aligned}
        &\mathcal{L}_{\mathrm{action}} = \mathrm{E}_{i,\tau,\epsilon'_i,a_i} \\
        &\quad \Bigl[\left\|u_{\phi}(a_{i,\tau},\tau\mid f_0,l)-(a_i-\epsilon'_i)\right\|_2^2\Bigr],
    \end{aligned}
    \label{eq:action-flow-matching-loss}
\end{equation}
where $u_{\phi}$ is the action velocity field parameterized by $\phi$. The Action Expert attends to the current-observation representations provided by the Observation Encoder, without directly receiving future-state or diff predictions. State and visual difference prediction influence action generation through joint training of the shared Observation Encoder.

\textbf{Overall Objective.} The complete training objective is a weighted combination of the three terms:
\begin{equation}
    \mathcal{L}=\mathcal{L}_{\mathrm{world}}+\lambda_{\mathrm{diff}}\mathcal{L}_{\mathrm{diff}}+\lambda_{\mathrm{action}}\mathcal{L}_{\mathrm{action}},
    \label{eq:overall-training-objective}
\end{equation}
where $\lambda_{\mathrm{diff}}$ and $\lambda_{\mathrm{action}}$ are hyperparameters balancing the three losses.

\subsection{Training Configurations and Inference}
\label{sec:training_inference}

\begin{figure*}[t]
    \centering
    \includegraphics[width=\textwidth]{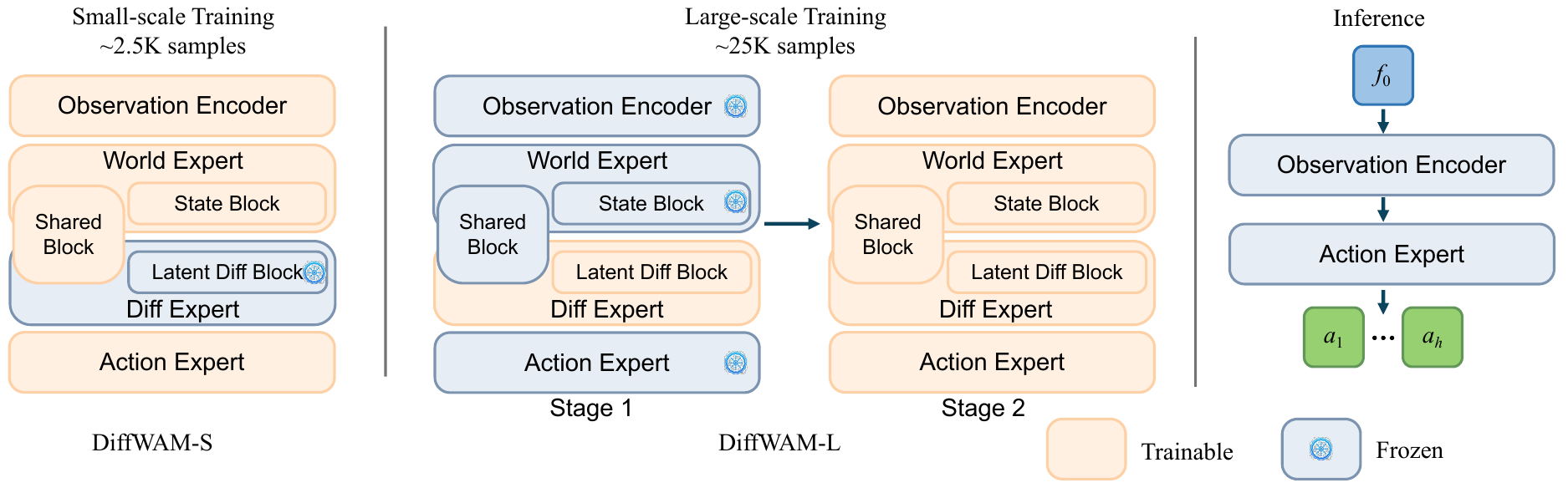}
    \caption{Training configurations and shared inference pathway of \modelname{}. \modelname{}-S freezes the independent LDB while retaining difference prediction during small-scale training. \modelname{}-L first trains the independent Diff Expert parameters with the remaining modules frozen, then jointly trains all modules. Orange denotes trainable modules, and blue with snowflake symbols denotes frozen modules; in \modelname{}-S, the shared World--Diff layers remain trainable. Both configurations retain only the Observation Encoder and Action Expert at inference.}
    \label{fig:train_inference}
\end{figure*}

We instantiate \modelname{} in two training configurations, \modelname{}-S for small-scale training and \modelname{}-L for large-scale training. Both configurations use the architecture and joint objective described above, with different parameter update schedules. Figure~\ref{fig:train_inference} illustrates these configurations and their shared inference pathway.

\textbf{\modelname{}-S.} For small-scale training, we freeze the independent $M$-layer LDB of the Diff Expert while jointly optimizing the Observation Encoder, World Expert, and Action Expert, including the visual layers shared by the World and Diff Experts. Freezing the LDB helps reduce overfitting to noise in small datasets while accelerating training by reducing the number of parameters to update. The diff loss remains active: its gradients propagate through the fixed LDB into the shared visual layers and Observation Encoder. Thus, difference prediction continues to shape the representation used for action generation even though the LDB parameters are frozen.

\textbf{\modelname{}-L.} For large-scale training, we adopt a two-stage procedure. In the first stage, the Observation Encoder, World Expert, and Action Expert are frozen, and only the independent Diff Expert parameters are optimized using the diff regression loss. The visual layers shared with the frozen World Expert also remain frozen. In the second stage, all modules are jointly optimized with the full objective, learning future-state prediction, difference prediction, and action generation together. This schedule allows the Diff Expert to first adapt to difference prediction with a fixed visual representation, before using the larger dataset to jointly refine the shared representation and all three experts.

\textbf{Inference.} Both configurations use the same action-generation pathway. The current observation is encoded by the Observation Encoder, and the Action Expert generates the action chunk conditioned on this representation and the language instruction. The World and Diff Experts are removed, so future-state and visual difference prediction introduce no additional inference overhead. The two configurations therefore differ in their training schedules while retaining the same inference architecture.

\section{Experiments}
\subsection{Implementation Details}

The Observation Encoder and the World, Diff, and Action Experts each contain $N$ Transformer layers. The World and Diff Experts use the Wan2.2-5B Transformer architecture~\cite{bib:wan}, with shared parameters in the first $N-M$ layers and separate parameters in the last $M$ layers. The Action Expert follows the same architecture but uses a reduced hidden dimension of $d_a=1024$. The complete model contains approximately 7B parameters. Further architectural, optimization, and inference details are provided in Appendix. The additional clean-to-random study starts task training without a task-trained FastWAM checkpoint. Separately, the continued-training control initializes both methods from the same task-trained FastWAM checkpoint and trains each for one additional epoch on clean demonstrations. We distinguish these two initialization protocols throughout the analysis.

\subsection{Experiment Setup}

We evaluate \modelname{} on LIBERO~\cite{bib:libero}, RoboTwin 2.0~\cite{bib:robotwin2}, and three real-world manipulation tasks on the ALOHA-AgileX platform. LIBERO covers 40 single-arm tasks across four suites, while RoboTwin 2.0 evaluates 50 bimanual tasks under clean and randomized conditions. We conduct 100 evaluation rollouts per task for LIBERO and the full-data RoboTwin 2.0 evaluation. For the additional clean-to-random evaluation, we evaluate each \modelname{} policy over 30 rollouts per task in each condition. For the real-world evaluation, we use approximately 1.1k demonstration trajectories in total across three tasks (Place Pens, Store Block, and Stack Cups), train for 63k optimization steps, and report the execution success rate over 20 trials per task for each method (60 trials per method in total). Main-table results are reported for individual training runs. Appendix~\ref{app:repeated_evaluation} additionally reports results from three independent training runs with different random seeds on LIBERO and the full-data and clean-only RoboTwin 2.0 settings. Each run produces a separately trained model that is evaluated using the corresponding benchmark protocol; we report the mean and sample standard deviation across these three runs to characterize variability across independently trained models. Complete benchmark and evaluation protocols are described in the appendix.

\subsection{Main Results}
This section evaluates \modelname{} on LIBERO, RoboTwin 2.0, and real-world manipulation tasks by comparing it with representative VLA and WAM methods and analyzing its different training configurations and training efficiency. Additional experimental results and detailed analyses are provided in the Appendix.

\subsubsection{Simulation Results}
\begin{table}[t]
    \centering
    {\resulttablefont
    \setlength{\tabcolsep}{1.5pt}
    \renewcommand{\arraystretch}{1.12}
\begin{tabular*}{\linewidth}{@{\extracolsep{\fill}}lccccc@{}}
        \toprule
        Method & Spatial & Object & Goal & Long & Avg. \\
        \midrule
        \multicolumn{6}{l}{$\mathit{VLA}$} \\
        OpenVLA~\cite{bib:openvla} & 84.70 & 88.40 & 79.20 & 53.70 & 76.50 \\
        OpenVLA-O~\cite{bib:openvla-oft} & 97.60 & 98.40 & 97.90 & 94.50 & 97.10 \\
        MemoryVLA~\cite{bib:memoryvla} & 98.40 & 98.40 & 96.40 & 93.40 & 96.65 \\
        $\pi$0.5~\cite{bib:pi05} & \underline{98.80} & 98.20 & 98.00 & 92.40 & 96.85 \\
        GROOT-N1~\cite{bib:groot-n1} & 94.40 & 97.60 & 93.00 & 90.60 & 93.90 \\
        \midrule
        \multicolumn{6}{l}{$\mathit{WAM}$} \\
        Motus~\cite{bib:motus} & 96.80 & \underline{99.80} & 96.60 & 97.60 & 97.70 \\
        Lingbot-VA~\cite{bib:lingbot-va} & 98.50 & 99.60 & 97.20 & \textbf{98.50} & 98.45 \\
        FastWAM$^{\dagger}$~\cite{bib:fastwam} & 98.20 & \underline{99.80} & 97.80 & 92.40 & 97.05 \\
        ImageWAM~\cite{bib:imagewam} & 97.20 & 99.20 & \underline{98.80} & \underline{98.40} & 98.40 \\
        WLA~\cite{bib:wla} & \textbf{99.00} & \textbf{100.00} & 97.80 & 97.60 & \underline{98.60} \\
        \midrule
        \modelname{}-S (Ours) & 98.00 & \underline{99.80} & \textbf{99.00} & \underline{98.40} & \textbf{98.80} \\
        \bottomrule
    \end{tabular*}

    }
    \caption{Main results on LIBERO. We use \modelname{}-S due to the small size of the LIBERO training dataset. $\dagger$ denotes our FastWAM baseline.}
    \label{tab:libero_main}
\end{table}

We evaluate \modelname{}-S on LIBERO and \modelname{}-L on RoboTwin 2.0, using the training configurations described in Section~\ref{sec:training_inference}, and compare them with representative VLA and WAM methods. The results are summarized in Table~\ref{tab:libero_main} and Table~\ref{tab:robotwin}.

For LIBERO, we report \modelname{}-S, the small-scale configuration described in Section~\ref{sec:training_inference}. It freezes the independent LDB parameters while retaining the diff loss and its gradient propagation into the shared backbone. As shown in Table~\ref{tab:libero_main}, \modelname{}-S achieves an average success rate of 98.8\%, improving upon FastWAM by 1.75 percentage points. The improvement is most pronounced on LIBERO-Goal and LIBERO-Long, where \modelname{}-S reaches 99.0\% and 98.4\%, outperforming FastWAM by 1.2 and 6.0 points, respectively. \modelname{}-S achieves the highest average success rate among the methods in the table, exceeding WLA (98.6\%) and LingBot-VA (98.45\%), with particularly clear gains over FastWAM on goal-conditioned and long-horizon tasks.

\begin{table}[t]
    \centering
    {\resulttablefont
    \setlength{\tabcolsep}{1.5pt}
    \renewcommand{\arraystretch}{1.18}
\begin{tabular*}{\linewidth}{@{\extracolsep{\fill}}lccc@{}}
        \toprule
        Method & Clean & Random & Avg. \\
        \midrule
        \multicolumn{4}{l}{$\mathit{VLA}$} \\
        $\pi$0~\cite{bib:pi0} & 65.92 & 58.80 & 62.36 \\
        X-VLA~\cite{bib:x-vla} & 72.08 & 72.84 & 72.46 \\
        $\pi$0.5~\cite{bib:pi05} & 82.74 & 76.76 & 79.75 \\
        ABot-M0~\cite{bib:abot-m0} & 86.60 & 85.08 & 85.84 \\
        QwenVLA~\cite{bib:qwen-vla} & 86.10 & 87.20 & 86.65 \\
        \midrule
        \multicolumn{4}{l}{$\mathit{WAM}$} \\
        Motus~\cite{bib:motus} & 88.66 & 87.02 & 87.84 \\
        Lingbot-VA~\cite{bib:lingbot-va} & 92.90 & \underline{91.50} & \underline{92.20} \\
        FastWAM$^{\dagger}$~\cite{bib:fastwam} & 91.88 & 90.32 & 91.10 \\
        X-WAM~\cite{bib:x-wam} & 89.76 & 90.68 & 90.22 \\
        WLA~\cite{bib:wla} & \underline{92.94} & 90.02 & 91.48 \\
        \midrule
        \modelname{}-L (Ours) & \textbf{93.28} & \textbf{92.02} & \textbf{92.65} \\
        \bottomrule
    \end{tabular*}

    }
    \caption{Main results on RoboTwin 2.0. We use \modelname{}-L due to the large size of the RoboTwin 2.0 training dataset. $\dagger$ denotes our FastWAM baseline.}
    \label{tab:robotwin}
\end{table}

For RoboTwin 2.0, we report \modelname{}-L, the large-scale configuration described in Section~\ref{sec:training_inference}. It first trains the independent LDB with the remaining modules frozen, then jointly optimizes all modules with the full objective. As shown in Table~\ref{tab:robotwin}, \modelname{}-L obtains an average success rate of 92.65\% on RoboTwin 2.0, outperforming the reproduced FastWAM baseline by 1.55 percentage points. The gain is consistent across both evaluation conditions, increasing performance from 91.88\% to 93.28\% in clean scenes and from 90.32\% to 92.02\% under scene randomization. \modelname{}-L also surpasses the strongest reported baselines in the table: its average success rate exceeds LingBot-VA (92.20\%) and WLA (91.48\%), while achieving the highest results in both the clean and randomized settings. Taken together, the results across both simulation benchmarks support our hypothesis that explicitly modeling consecutive visual differences yields more task-relevant representations for complex manipulation.
\subsubsection{Clean-to-Random Generalization}
\begin{table}[t]
    \centering
    {\resulttablefont
    \setlength{\tabcolsep}{1.5pt}
    \renewcommand{\arraystretch}{1.12}
\begin{tabular*}{\linewidth}{@{\extracolsep{\fill}}lcccc@{}}
        \toprule
        Method & Embodied PT & Clean & Random & Avg. \\
        \midrule
        \multicolumn{5}{@{}l}{\textit{With Embodied Pretraining}} \\
        $\pi_0$~\cite{bib:pi0} & Yes & 46.4 & 16.3 & 31.35 \\
        X-VLA~\cite{bib:x-vla} & Yes & 68.0 & 20.9 & 44.45 \\
        $\pi_{0.5}$~\cite{bib:pi05} & Yes & 70.7 & 46.0 & 58.35 \\
        \midrule
        \multicolumn{5}{@{}l}{\textit{Without Embodied Pretraining}} \\
        StarVLA~\cite{bib:starvla} & No & 46.5 & 3.2 & 24.85 \\
        AHA-WAM~\cite{bib:aha-wam} & No & 64.3 & 3.2 & 33.75 \\
        FastWAM$^{\dagger}$~\cite{bib:fastwam} & No & \textbf{70.2} & 1.50 & 35.85 \\
        \modelname{} (Ours) & No & 69.7 & \textbf{5.27} & \textbf{37.47} \\
        \bottomrule
    \end{tabular*}

    }
    \caption{\textbf{Clean-to-random generalization on RoboTwin 2.0.} Success rates (\%) for policies trained on clean demonstrations and evaluated in clean and randomized environments without further adaptation. $\dagger$ denotes our FastWAM baseline.}
    \label{tab:robotwin_clean_random}
\end{table}

To evaluate transfer beyond the training distribution, we train on clean RoboTwin 2.0 demonstrations and directly evaluate under randomized conditions without further adaptation. Unlike the full-data experiment in Table~\ref{tab:robotwin}, randomized demonstrations are excluded from these task-training runs. This study starts task training without a task-trained FastWAM checkpoint and is separate from the continued-training control below. As shown in Table~\ref{tab:robotwin_clean_random}, \modelname{} achieves 5.27\% success, compared with 1.50\% for our FastWAM baseline, an absolute gain of 3.77 percentage points. Clean success decreases slightly from 70.2\% to 69.7\%, while the mean of Clean and Random improves from 35.85\% to 37.47\%. It also exceeds the 3.2\% reported for StarVLA and AHA-WAM among the selected methods without embodied pretraining. These results show improved transfer under the evaluated RoboTwin distribution shift, with a small trade-off in clean performance. External results provide context, while our FastWAM evaluation is the directly compared baseline.

\subsubsection{Real-World Results}
\begin{figure*}[!t]
    \centering
    \includegraphics[width=\linewidth]{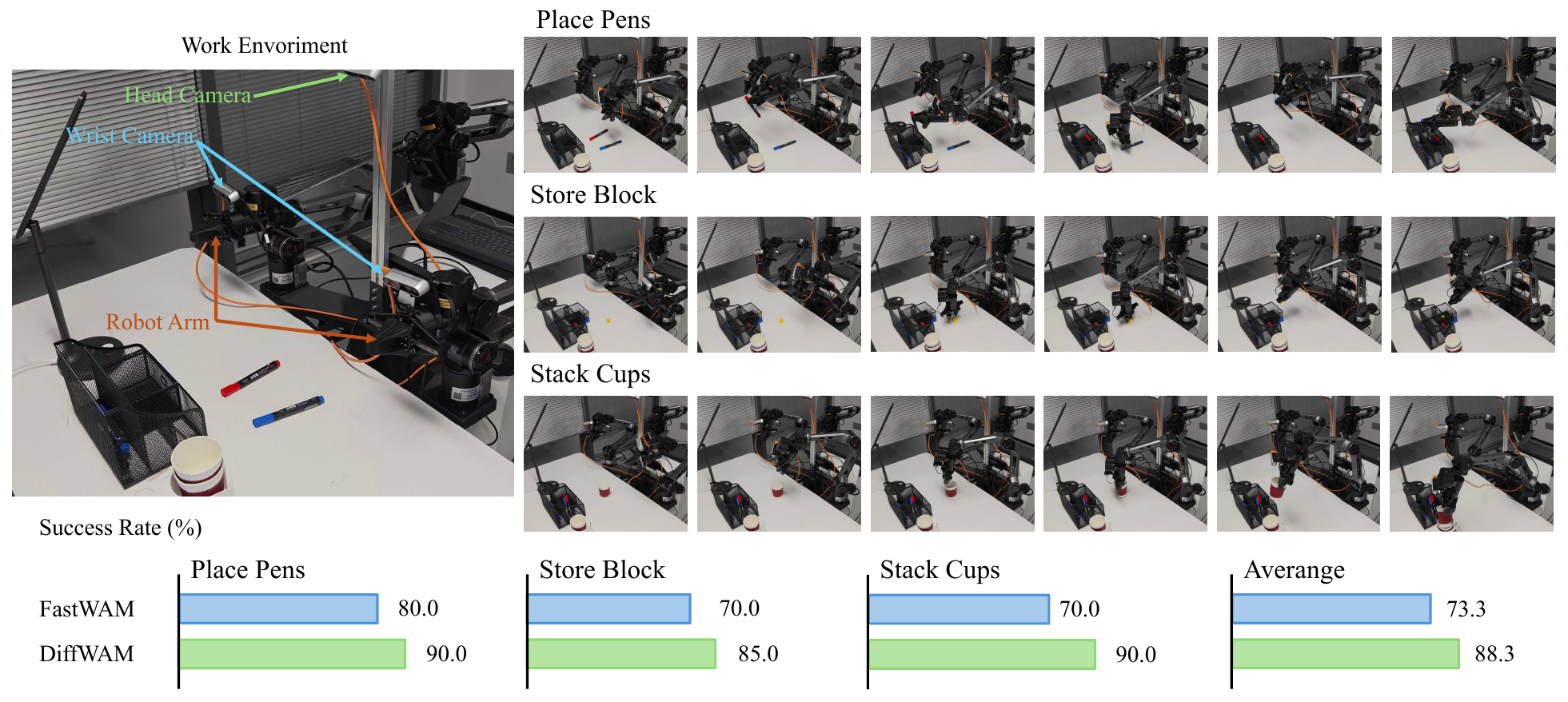}
    \caption{Real-world setup, task execution sequences, and success rates. Left: the robot workspace and camera placement. Right: example executions of Place Pens, Store Block, and Stack Cups. Bottom: success rates of FastWAM and \modelname{}-S across the three tasks.}
    \label{fig:real_world}
\end{figure*}

We use the \modelname{}-S configuration for real-world evaluation on three manipulation tasks: Place Pens, Store Block, and Stack Cups. The training dataset contains approximately 1.1k demonstration trajectories in total, and the model is trained for 63k optimization steps. Each method is evaluated over 20 trials per task, totaling 60 trials across the three tasks. Figure~\ref{fig:real_world} shows the robot setup, task execution sequences, and success rates. \modelname{}-S achieves success rates of 90\%, 85\%, and 90\% on the three tasks, respectively, compared with 80\%, 70\%, and 70\% for FastWAM. Averaged across the three tasks, \modelname{}-S achieves 88.3\% success versus 73.3\% for FastWAM, an absolute improvement of 15.0 percentage points. These results support the benefits of explicitly modeling consecutive visual differences across real-world manipulation tasks involving different objects and motions.

\subsubsection{Comparison of Different Configurations}
\begin{table}[t]
    \centering
    {\resulttablefont
    \setlength{\tabcolsep}{1.5pt}
    \renewcommand{\arraystretch}{1.12}
\begin{tabular*}{\linewidth}{@{\extracolsep{\fill}}lccc@{}}
        \toprule
        Setting & FastWAM & \modelname{}-L & \modelname{}-S \\
        \midrule
        LIBERO & 97.05 & 97.70 & \textbf{98.80} \\
        RoboTwin-Clean & 88.47 & 87.72 & \textbf{93.13} \\
        RoboTwin-Full & 91.88 & \textbf{93.28} & 92.60 \\
        \bottomrule
    \end{tabular*}

    }
    \caption{Success rates (\%) of \modelname{}-L and \modelname{}-S. \modelname{}-L updates the LDB together with the remaining modules during its joint training stage; \modelname{}-S freezes the independent LDB parameters while retaining the diff loss. RoboTwin-Clean and RoboTwin-Full use clean-scene and full training data, respectively; both are evaluated in the clean setting.}
    \label{tab:small_transfer}
\end{table}

We compare \modelname{}-L and \modelname{}-S across different training data scales. We consider two data regimes: small-scale training on LIBERO and RoboTwin-Clean, with approximately 2k and 2.5k demonstration trajectories, respectively, and large-scale training on RoboTwin-Full, with approximately 27k demonstration trajectories.

As shown in Table~\ref{tab:small_transfer}, \modelname{}-L achieves 97.70\% success on LIBERO, compared with 97.05\% for FastWAM, while \modelname{}-S reaches 98.80\%. On RoboTwin-Clean, \modelname{}-L obtains 87.72\%, below the FastWAM baseline of 88.47\%, whereas \modelname{}-S achieves 93.13\%. \modelname{}-S therefore outperforms \modelname{}-L by 1.10 and 5.41 percentage points in these two settings. With the full RoboTwin training set, \modelname{}-L instead achieves 93.28\%, exceeding \modelname{}-S's 92.60\% by 0.68 percentage points; both outperform FastWAM's 91.88\%.

\modelname{}-S constrains optimization by freezing the independent difference-prediction module, reducing the number of parameters fitted to limited demonstrations while retaining difference supervision for the shared representation. This configuration balances adaptation with regularization, addressing the risk of overfitting to noise in small datasets. \modelname{}-L, in contrast, jointly optimizes all modules, providing greater flexibility to learn from the more diverse trajectories in the larger RoboTwin-Full training set. The performance comparison reflects this trade-off: \modelname{}-S achieves higher success rates in the two small-data settings, whereas \modelname{}-L performs better in the large-data setting. These results support choosing \modelname{}-S when training data are limited and \modelname{}-L when more data are available. Task-level results for full-data RoboTwin 2.0 training are provided in the appendix.

\subsubsection{Training Efficiency}
\begin{figure}[t]
    \centering
    \includegraphics[width=\columnwidth]{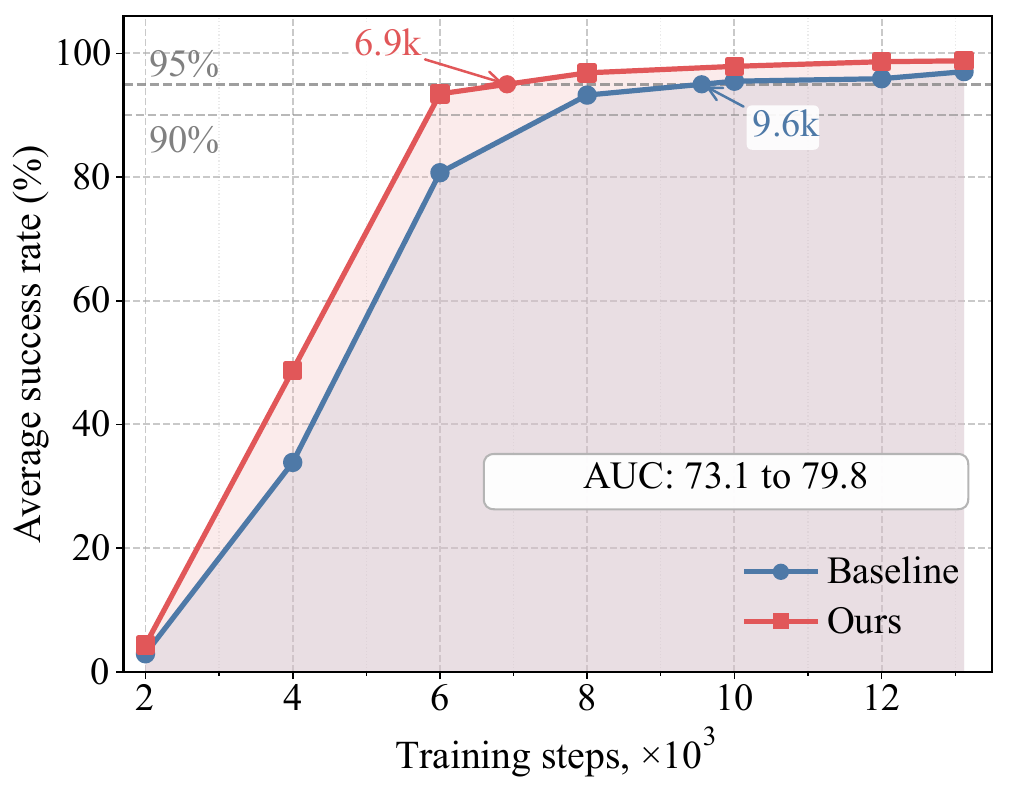}
    \caption{Training performance comparison on LIBERO.}
    \label{fig:training_performance}
    \vspace{-0.3cm}
\end{figure}

Figure~\ref{fig:training_performance} compares the training dynamics of \modelname{} and FastWAM (baseline) on LIBERO. As shown in Figure~\ref{fig:training_performance}, \modelname{} improves substantially faster during the early stage of training and reaches a 95\% average success rate after only 6.9k steps, whereas FastWAM requires 9.6k steps. This corresponds to a 28.1\% reduction in the number of optimization steps needed to reach the same performance level. \modelname{} also increases the convergence AUC from 73.1 to 79.8 and achieves a higher final success rate. These results indicate that visual difference supervision provides a more informative learning signal, enabling the policy to acquire task-relevant representations with fewer optimization steps while also improving its asymptotic performance.

\subsection{Ablation Studies}
\subsubsection{Architecture Ablation}
\begin{table}[!t]
    \centering
    {\resulttablefont
    \setlength{\tabcolsep}{1.5pt}
    \renewcommand{\arraystretch}{1.12}
\begin{tabular*}{\linewidth}{@{\extracolsep{\fill}}lccccc@{}}
        \toprule
        Architecture & Spatial & Object & Goal & Long & Avg. \\
        \midrule
        World Expert & \textbf{98.20} & 99.80 & 97.80 & 92.40 & 97.05 \\
        \shortstack[l]{World Expert\\+ Diff Loss} & 97.80 & \textbf{100.00} & 97.40 & 95.80 & 97.75 \\
        \shortstack[l]{World Expert\\+ Diff Expert} & 98.00 & 99.80 & \textbf{99.00} & \textbf{98.40} & \textbf{98.80} \\
        \bottomrule
    \end{tabular*}

    }
    \caption{Architecture ablation on LIBERO. World Expert uses the measured FastWAM results, World Expert + Diff Loss adds difference supervision without a separate Diff Expert, and World Expert + Diff Expert uses the \modelname{}-S results. The Observation Encoder and Action Expert are retained in all configurations.}
    \label{tab:architecture_ablation}
\end{table}

Table~\ref{tab:architecture_ablation} examines whether the gains of \modelname{} arise from difference supervision alone or also benefit from a dedicated Diff Expert. Starting from the World Expert baseline (FastWAM), adding the diff loss raises average success from 97.05\% to 97.75\%. The clearest gain is on LIBERO-Long, where success increases from 92.40\% to 95.80\% (+3.40 points), suggesting that explicit difference supervision is particularly useful for tasks requiring sustained execution over multiple steps. However, loss-only supervision does not improve every suite: Spatial and Goal decrease slightly, while Object reaches 100.00\%.

Retaining the diff loss and introducing the separate Diff Expert further raises average success to 98.80\% (+1.05 points over loss-only supervision). This additional gain is concentrated on Goal and Long, which improve by 1.60 and 2.60 points, respectively, while Spatial recovers by 0.20 points and Object changes only slightly. Relative to FastWAM, the full architecture improves Long by 6.00 points and Goal by 1.20 points, indicating that its advantage is most pronounced in long-horizon execution and goal-conditioned manipulation. These results suggest that a dedicated difference-prediction pathway helps realize the benefits of difference supervision beyond simply adding an auxiliary loss. Thus, the diff loss provides a measurable improvement, and the Diff Expert contributes a further gain; both are necessary to attain the full performance of \modelname{} in this ablation.

\subsubsection{Number of LDB Layers}
\begin{table}[!t]
    \centering
    {\resulttablefont
    \setlength{\tabcolsep}{1.5pt}
    \renewcommand{\arraystretch}{1.12}
\begin{tabular*}{\linewidth}{@{\extracolsep{\fill}}lccccc@{}}
        \toprule
        Method & Spatial & Object & Goal & Long & Avg. \\
        \midrule
        FastWAM~\cite{bib:fastwam} & 98.20 & 99.80 & 97.80 & 92.40 & 97.05 \\
        \modelname{} ($M{=}5$) & \textbf{99.60} & \textbf{100.00} & 97.20 & 96.00 & 98.20 \\
        \modelname{} ($M{=}10$) & 98.00 & 99.80 & \textbf{99.00} & \textbf{98.40} & \textbf{98.80} \\
        \modelname{} ($M{=}15$) & 98.20 & 99.80 & 97.40 & 97.00 & 98.10 \\
        \modelname{} ($M{=}20$) & 97.00 & \textbf{100.00} & 98.40 & 98.00 & 98.35 \\
        \bottomrule
    \end{tabular*}

    }
    \caption{Ablation study of LDB layers $M$ on LIBERO. All \modelname{} entries use the \modelname{}-S configuration.}
    \label{tab:m_ablation}
\end{table}

We first investigate the effect of the LDB depth $M$ on LIBERO. All \modelname{} entries in Table~\ref{tab:m_ablation} use the \modelname{}-S configuration, with the independent LDB parameters frozen and the diff loss retained. All tested LDB depths outperform the FastWAM baseline on average. Increasing $M$ from 5 to 10 improves the average success rate from 98.20\% to 98.80\%, with particularly strong gains on LIBERO-Goal and LIBERO-Long. Further increasing the depth to 15 or 20 layers does not yield additional improvements. These results favor a moderate depth for the difference-prediction pathway; we therefore use $M=10$ as the default configuration.

\subsubsection{Visual Expert Configurations}
\begin{table}[!t]
    \centering
    {\resulttablefont
    \setlength{\tabcolsep}{1.5pt}
    \renewcommand{\arraystretch}{1.12}
\begin{tabular*}{\linewidth}{@{\extracolsep{\fill}}lccccc@{}}
        \toprule
        Visual Experts & Spatial & Object & Goal & Long & Avg. \\
        \midrule
        Only World Expert & 98.20 & 99.80 & 97.80 & 92.40 & 97.05 \\
        Only Diff Expert & \textbf{99.20} & \textbf{100.00} & 95.80 & 89.80 & 96.20 \\
        World + Diff Expert & 98.00 & 99.80 & \textbf{99.00} & \textbf{98.40} & \textbf{98.80} \\
        \bottomrule
    \end{tabular*}

    }
    \caption{Ablation study of visual expert configurations on LIBERO.}
    \label{tab:loss_ablation}
\end{table}

We next compare the visual expert configurations used during training. Table~\ref{tab:loss_ablation} considers Only World Expert, Only Diff Expert, and World + Diff Expert. These names refer to the visual prediction pathways; the Observation Encoder and Action Expert are retained for action learning in all three configurations. Only World Expert learns future-state prediction, Only Diff Expert learns latent-difference prediction, and World + Diff Expert jointly learns both.

As shown in Table~\ref{tab:loss_ablation}, Only World Expert uses the measured FastWAM baseline and achieves an average success rate of 97.05\%, while Only Diff Expert obtains 96.20\%. World + Diff Expert improves the average success rate to 98.80\%, outperforming the two single-expert configurations by 1.75 and 2.60 percentage points, respectively. The improvement is especially pronounced on LIBERO-Long, where the combined configuration reaches 98.40\%, compared with 92.40\% for Only World Expert and 89.80\% for Only Diff Expert.

These results support the complementary roles of the two visual experts: the World Expert models future scene states, while the Diff Expert explicitly captures inter-frame visual differences. Jointly learning these capabilities enhances the representation used for action generation, particularly on long-horizon tasks.

\subsubsection{Controlling for Additional Training}
\begin{table}[t]
    \centering
    {\small
    \setlength{\tabcolsep}{7pt}
    \renewcommand{\arraystretch}{1.12}
    \begin{adjustbox}{max width=\linewidth}
\begin{tabular}{@{}lcc@{}}
        \toprule
        Method & Additional training & Clean success (\%) \\
        \midrule
        FastWAM (continued training) & 1 epoch & 88.47 \\
        \modelname{} (Ours) & 1 epoch & \textbf{93.13} \\
        \bottomrule
    \end{tabular}
\end{adjustbox}
    }
    \caption{\textbf{Controlling for additional training on RoboTwin 2.0.} Both methods start from the same task-trained FastWAM checkpoint and receive one additional epoch of training on clean demonstrations under matched conditions. This experiment is separate from the clean-to-random study in Table~\ref{tab:robotwin_clean_random}.}
    \label{tab:continued_training}
\end{table}

To distinguish the effect of difference supervision from that of additional optimization, we initialize FastWAM and \modelname{} from the same task-trained FastWAM checkpoint and train both for one additional epoch on clean demonstrations under matched training conditions. As shown in Table~\ref{tab:continued_training}, \modelname{} achieves 93.13\% clean success, compared with 88.47\% for continued FastWAM training, a gain of 4.66 percentage points. This comparison supports an improvement beyond additional training alone and complements the visual-expert ablation in Table~\ref{tab:loss_ablation}. The shared-checkpoint initialization isolates the effect of the modified training objective under a matched additional training budget.

\subsection{Visualization}
\begin{figure}[t]
    \centering
    \includegraphics[width=\columnwidth]{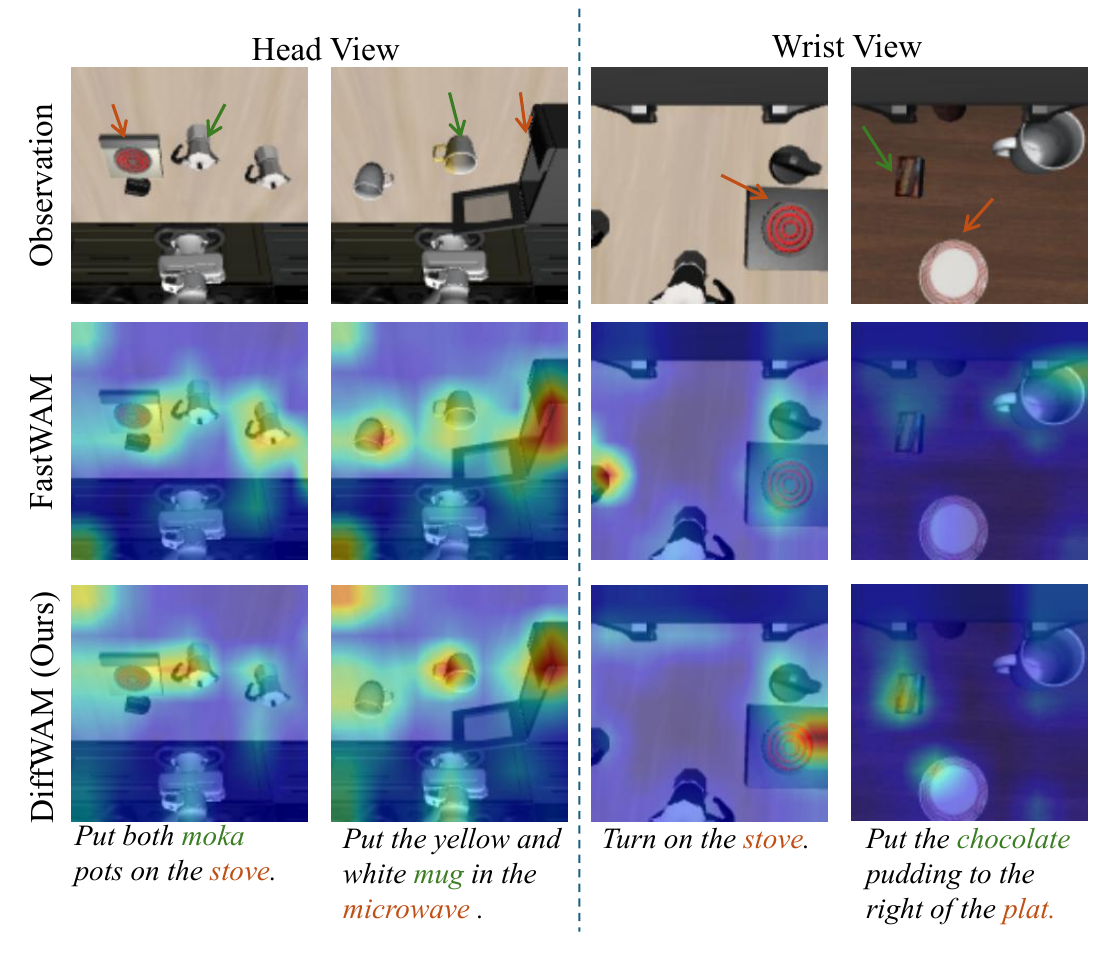}
    \caption{Attention visualizations. Arrows indicate target regions specified by the language instruction. 
    }
    \label{fig:attention_small}
    \vspace{-0.4cm}
\end{figure}

Figure~\ref{fig:attention_small} visualizes the attention responses of \modelname{} from the head and wrist camera views. The attention maps retain fine-grained scene texture, while assigning stronger responses to the objects being manipulated. This indicates that \modelname{} can preserve detailed visual context without dispersing its representation uniformly over the scene; instead, it selectively concentrates attention on action-relevant objects. Overall, the visualization supports our design goal of learning representations that balance rich scene understanding with focused perception of the manipulated object.

\FloatBarrier
\section{Conclusion}
In this work, we proposed \modelname{}, a World Action Model that enhances world representations by jointly learning future-state prediction, visual difference prediction, and action generation. Its Mixture-of-Experts architecture combines a shared Observation Encoder with dedicated World, Diff, and Action Experts. A structured unidirectional attention mask allows each expert to access the current-observation representation while separating their prediction pathways. At inference, the World and Diff Experts are removed, retaining only the Observation Encoder and Action Expert without additional inference overhead.

Experiments on LIBERO, RoboTwin 2.0, and real-world manipulation demonstrate the effectiveness of joint state and visual difference modeling. \modelname{}-S achieves average success rates of 98.8\% on LIBERO and 88.3\% on the real-world tasks, while \modelname{}-L achieves 92.65\% on RoboTwin 2.0. Comparisons across data settings favor \modelname{}-S for smaller training sets and \modelname{}-L for the full RoboTwin training set, supporting the choice of training configuration according to data availability. An additional clean-to-random evaluation on RoboTwin 2.0 improves success from 1.50\% to 5.27\%, and a separate matched checkpoint-continuation control improves clean success from 88.47\% to 93.13\%. Ablation studies further confirm the complementary roles of the World and Diff Experts. Future work will explore large-scale pretraining and extend state and visual difference modeling to more diverse manipulation settings.

\nocite{*}
\bibliographystyle{unsrtnat}
\bibliography{main}

\clearpage
\appendix

\section{Additional Experimental Details}
\label{app:experimental_details}

The key configurations for the three experimental settings are summarized in Table~\ref{tab:implementation_configs}. We use \modelname{}-S for the main LIBERO and real-world experiments and \modelname{}-L for the main RoboTwin 2.0 experiments.
\begin{table}[!htbp]
    \centering
    {\small
    \setlength{\tabcolsep}{2pt}
    \renewcommand{\arraystretch}{1.04}
    \begin{adjustbox}{max width=\linewidth}
\begin{tabular}{lccc}
        \toprule
        Setting & LIBERO & RoboTwin 2.0 & Real \\
        \midrule
        Model configuration & \modelname{}-S & \modelname{}-L & \modelname{}-S \\
        Training hardware & $8\times$ A100 & $8\times$ A100 & $8\times$ A100 \\
        Grad. accumulation & 2 & 2 & 2 \\
        Weight decay & $1\times10^{-2}$ & $1\times10^{-2}$ & $1\times10^{-2}$ \\
        LR schedule & Cosine & Cosine & Cosine \\
        Learning rate & $1\times10^{-4}$ & $1\times10^{-5}$ & $1\times10^{-4}$ \\
        Batch size & 8 & 8 & 8 \\
        Training steps & $1.3\times10^4$ & $3\times10^4$ & $6.3\times10^4$ \\
        Network depth $N$ & 30 & 30 & 30 \\
        Network depth $M$ & 10 & 10 & 10 \\
        Cameras & 2 & 3 & 2 \\
        Video size & $[224,448]$ & $[384,320]$ & $[224,576]$ \\
        Image resize & $[224,224]$ & $[240,320]$ & $[240,320]$ \\
        Action dim. & 7 & 14 & 7 \\
        Proprio dim. & 8 & 14 & 7 \\
        Normalization & min/max & z-score & z-score \\
        Validation split & 0.0 & 0.01 & 0.0 \\
        \bottomrule
    \end{tabular}
\end{adjustbox}
    }
    \caption{Implementation configurations for each experimental setting.}
    \label{tab:implementation_configs}
\end{table}

\subsubsection{LIBERO} LIBERO~\cite{bib:libero} consists of four suites: LIBERO-Spatial, LIBERO-Object, LIBERO-Goal, and LIBERO-Long. Each suite contains 10 tasks and provides 500 expert demonstrations, resulting in 40 tasks and 2,000 demonstrations in total. We conduct 100 rollouts per task, amounting to 4,000 evaluation trials, and report the average success rate for each suite and across all four suites. For this benchmark, \modelname{}-S is initialized from Wan2.2-5B and trained for $1.3\times10^4$ steps with a learning rate of $1\times10^{-4}$ and a batch size of 8. The model uses two camera views, a 7-dimensional action space, an 8-dimensional proprioceptive input, min/max normalization, and network depths of $N=30$ and $M=10$.

\subsubsection{RoboTwin 2.0} RoboTwin 2.0~\cite{bib:robotwin2} evaluates bimanual manipulation under both clean and heavily randomized environments. We adopt its multi-task setting with 50 tasks and use approximately 27k demonstration trajectories in total, comprising 2,500 demonstrations from clean scenes and 25,000 demonstrations with strong scene randomization. Each task is evaluated over 100 trials under both clean and randomized conditions, and we report the success rates for the two settings and their average. Due to computational resource constraints, \modelname{}-L is initialized from FastWAM for this benchmark and trained for $3\times10^4$ steps with a learning rate of $1\times10^{-5}$ and a batch size of 8. The model uses three camera views, a 14-dimensional action space, a 14-dimensional proprioceptive input, z-score normalization, and network depths of $N=30$ and $M=10$.

\ifdefined\standalonesupplement
\begin{table}[t]
    \centering
    {\resulttablefont
    \setlength{\tabcolsep}{1.5pt}
    \renewcommand{\arraystretch}{1.12}
\begin{tabular*}{\linewidth}{@{\extracolsep{\fill}}lcccc@{}}
        \toprule
        Method & Embodied PT & Clean & Random & Avg. \\
        \midrule
        \multicolumn{5}{@{}l}{\textit{With Embodied Pretraining}} \\
        $\pi_0$~\cite{bib:pi0} & Yes & 46.4 & 16.3 & 31.35 \\
        X-VLA~\cite{bib:x-vla} & Yes & 68.0 & 20.9 & 44.45 \\
        $\pi_{0.5}$~\cite{bib:pi05} & Yes & 70.7 & 46.0 & 58.35 \\
        \midrule
        \multicolumn{5}{@{}l}{\textit{Without Embodied Pretraining}} \\
        StarVLA~\cite{bib:starvla} & No & 46.5 & 3.2 & 24.85 \\
        AHA-WAM~\cite{bib:aha-wam} & No & 64.3 & 3.2 & 33.75 \\
        FastWAM$^{\dagger}$~\cite{bib:fastwam} & No & \textbf{70.2} & 1.50 & 35.85 \\
        \modelname{} (Ours) & No & 69.7 & \textbf{5.27} & \textbf{37.47} \\
        \bottomrule
    \end{tabular*}

    }
    \caption{\textbf{Clean-to-random generalization on RoboTwin 2.0.} Success rates (\%) for policies trained on clean demonstrations and evaluated in clean and randomized environments without further adaptation. $\dagger$ denotes our FastWAM baseline.}
    \label{tab:robotwin_clean_random}
\end{table}

\begin{table}[t]
    \centering
    {\small
    \setlength{\tabcolsep}{7pt}
    \renewcommand{\arraystretch}{1.12}
    \begin{adjustbox}{max width=\linewidth}
\begin{tabular}{@{}lcc@{}}
        \toprule
        Method & Additional training & Clean success (\%) \\
        \midrule
        FastWAM (continued training) & 1 epoch & 88.47 \\
        \modelname{} (Ours) & 1 epoch & \textbf{93.13} \\
        \bottomrule
    \end{tabular}
\end{adjustbox}
    }
    \caption{\textbf{Controlling for additional training on RoboTwin 2.0.} Both methods start from the same task-trained FastWAM checkpoint and receive one additional epoch of training on clean demonstrations under matched conditions. This experiment is separate from the clean-to-random study in Table~\ref{tab:robotwin_clean_random}.}
    \label{tab:continued_training}
\end{table}

\fi

\subsubsection{Clean-to-random evaluation}
The additional experiment in Table~\ref{tab:robotwin_clean_random} uses clean demonstrations only during task training and evaluates the resulting policies under randomized RoboTwin 2.0 conditions without adaptation. These runs start task training without loading a task-trained FastWAM checkpoint; training from scratch here refers to robot-task training and does not imply that all model parameters are randomly initialized. This experiment is distinct from the full-data training configuration above, which includes randomized demonstrations. Clean/Random success is 70.2\%/1.50\% for our FastWAM baseline and 69.7\%/5.27\% for \modelname{}. We evaluate \modelname{} over 30 trials per task in each condition, totaling 1,500 trials per condition. It succeeds in 1,045 clean trials and 79 randomized trials, yielding unrounded success rates of 1,045/1,500 and 79/1,500. The Clean/Random average is computed before rounding as 1,124/3,000, or 37.47\% when rounded to two decimal places. The corresponding FastWAM average is 35.85\%. Clean results from the separate checkpoint-continuation experiment are not paired with these values.

\subsubsection{Checkpoint-continuation control}
For Table~\ref{tab:continued_training}, FastWAM and \modelname{} instead share the same task-trained FastWAM checkpoint as their starting point, with the additional Diff pathway introduced for \modelname{}. Both receive one additional epoch of training on clean demonstrations under matched conditions and are evaluated in the clean setting. The reported success rates are 88.47\% and 93.13\%, respectively. This comparison controls for the additional training budget and is separate from both the clean-to-random evaluation and the full-data RoboTwin result. No mechanistic claim about overfitting is made from the FastWAM score alone.

\subsubsection{Real-world} We evaluate \modelname{}-S on the ALOHA-AgileX platform,\footnote{\url{https://github.com/agilexrobotics/aloha-agilex}} using approximately 1.1k demonstration trajectories in total across Place Pens, Store Block, and Stack Cups and reporting the execution success rate over 20 trials per task for each method (60 trials per method in total). For these experiments, \modelname{}-S is initialized from Wan2.2-5B and trained for $6.3\times10^4$ steps with a learning rate of $1\times10^{-4}$ and a batch size of 8. The model uses two camera views, a 7-dimensional action space, a 7-dimensional proprioceptive input, z-score normalization, and network depths of $N=30$ and $M=10$.

\section{Additional Experimental Results}
\label{app:additional_results}

This section describes the comparison methods and provides task-level RoboTwin 2.0 results, real-world analyses, visualizations of state and visual difference prediction, and supplementary training-performance results.

\subsection{Comparison Methods}

We compare \modelname{} with representative vision language action (VLA) and world action model (WAM) approaches. VLA methods directly map visual-language observations to actions, while WAM methods additionally model action-conditioned dynamics.

\subsubsection{VLAs} OpenVLA~\cite{bib:openvla} and OpenVLA-O~\cite{bib:openvla-oft} are an open VLA baseline and its optimized variant; MemoryVLA~\cite{bib:memoryvla} augments VLA control with memory. $\pi$0~\cite{bib:pi0} and $\pi$0.5~\cite{bib:pi05} are flow-based VLA models, with the latter targeting open-world generalization. GR00T-N1~\cite{bib:groot-n1} is a generalist humanoid foundation model, X-VLA~\cite{bib:x-vla} targets cross-embodiment transfer, ABot-M0~\cite{bib:abot-m0} learns an action manifold, and QwenVLA~\cite{bib:qwen-vla} unifies VLA modeling across tasks and embodiments.

\subsubsection{WAMs} Motus~\cite{bib:motus} models latent actions in a unified world model; LingBot-VA~\cite{bib:lingbot-va} uses causal world modeling; and FastWAM~\cite{bib:fastwam} is the WAM baseline used for RoboTwin 2.0 initialization. ImageWAM~\cite{bib:imagewam}, X-WAM~\cite{bib:x-wam}, and WLA~\cite{bib:wla} respectively explore image editing based WAM, 4D world action modeling, and joint world language action modeling.

\subsection{RoboTwin 2.0 Results}
\subsubsection{Full data training} Table~\ref{tab:robotwin_task_full} reports task-level results when training with the full RoboTwin 2.0 data. \modelname{}-L achieves 93.28\% success in clean scenes and 92.02\% under randomization, for an average of 92.65\%. This improves over the reproduced FastWAM baseline by 1.55 percentage points overall, with gains in both clean and randomized conditions (91.88\% to 93.28\% and 90.32\% to 92.02\%, respectively). At the task level, \modelname{}-L is competitive across a broad range of manipulation skills and reaches or ties the best result on many tasks, indicating that explicitly modeling inter-frame visual differences improves performance without sacrificing robustness to scene variation.

\subsection{Independent Training Runs and Uncertainty}
\label{app:repeated_evaluation}
For each of LIBERO and the full-data and clean-only RoboTwin 2.0 settings, we independently train our method three times with different random seeds and evaluate each resulting model using the corresponding benchmark protocol. Table~\ref{tab:repeated_evaluation} reports the score from each training run, together with the mean and sample standard deviation across the three runs. The resulting scores are $92.52\pm0.11$\% on full-data RoboTwin 2.0, $98.67\pm0.12$\% on LIBERO, and $37.40\pm0.07$\% for the Clean/Random average in the clean-only RoboTwin setting.

\begin{table}[!t]
\centering
{\resulttablefont
\setlength{\tabcolsep}{1.5pt}
\renewcommand{\arraystretch}{1.15}
\begin{tabular*}{\linewidth}{@{\extracolsep{\fill}}lcccc@{}}
\toprule
Setting & Run 1 & Run 2 & Run 3 & Mean $\pm$ Std. \\
\midrule
\shortstack[l]{RoboTwin 2.0\\Full-data} & 92.49 & 92.65 & 92.43 & $92.52 \pm 0.11$ \\
LIBERO & 98.60 & 98.80 & 98.60 & $98.67 \pm 0.12$ \\
\shortstack[l]{RoboTwin 2.0\\Clean-only} & 37.33 & 37.40 & 37.47 & $37.40 \pm 0.07$ \\
\bottomrule
\end{tabular*}}
\caption{\textbf{Independent training runs of \modelname{}.} Success rates (\%) for three independently trained and evaluated models; mean $\pm$ sample standard deviation ($n=3$, denominator $n-1$). RoboTwin scores average Clean and Random success; full-data and clean-only specify training settings. The clean-only average is not Random-only OOD success. LIBERO averages its four suites.}
\label{tab:repeated_evaluation}
\end{table}

The reported metrics exhibit limited variability across the three independently trained models, supporting the consistency of the method across training seeds in these settings. Each observation corresponds to a separate training run followed by evaluation, so the standard deviations summarize variation across complete training-and-evaluation runs. Main-table scores correspond to individual runs and are reported separately from the three-run means. In particular, the clean-only RoboTwin metric averages Clean and Random success and should not be interpreted as a Random-only success rate.

\subsection{Real-world Results}
\ifdefined\standalonesupplement
\begin{figure*}[!t]
    \centering
    \includegraphics[width=\linewidth]{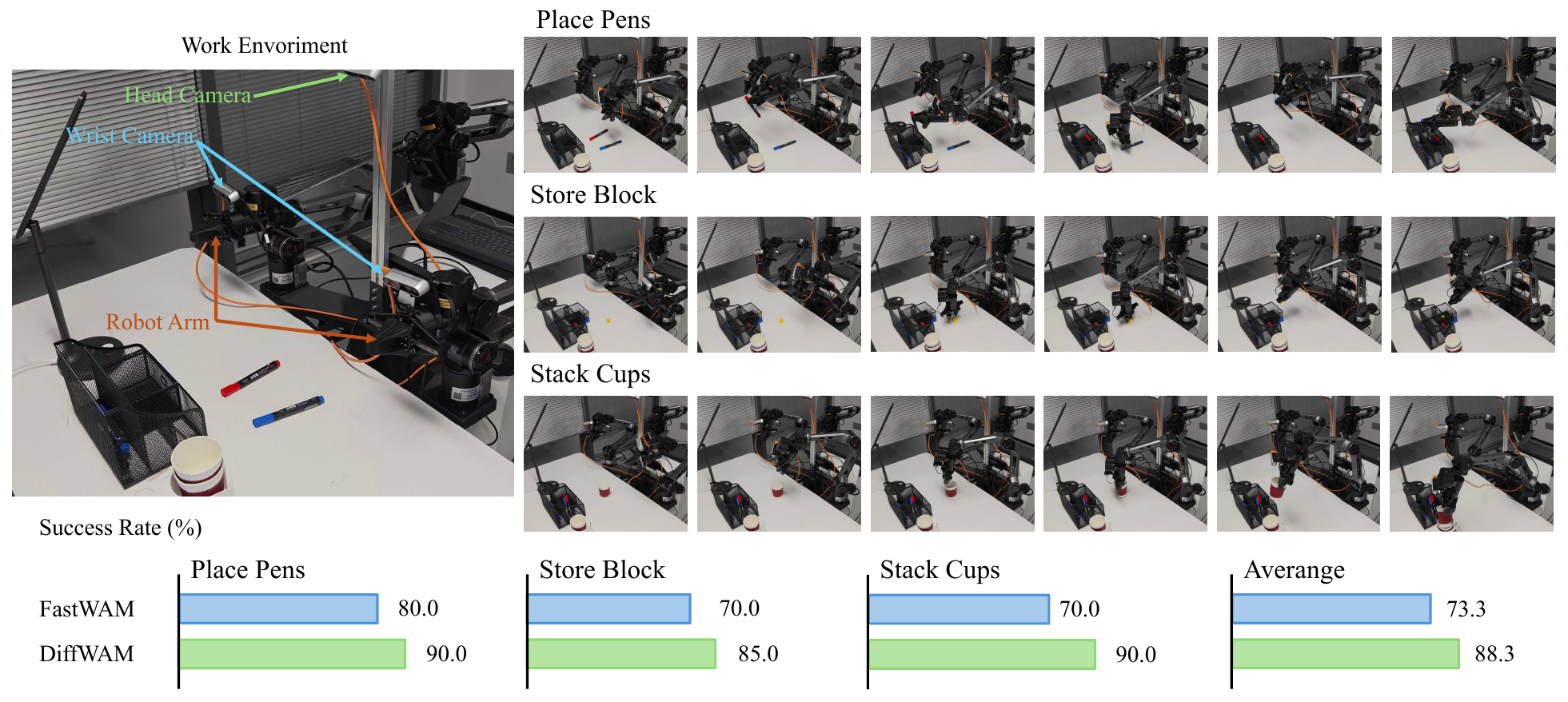}
    \caption{Real-world setup, task execution sequences, and success rates. Left: the robot workspace and camera placement. Right: example executions of Place Pens, Store Block, and Stack Cups. Bottom: success rates of FastWAM and \modelname{}-S across the three tasks.}
    \label{fig:real_world}
\end{figure*}

\fi
Our real-world study evaluates \modelname{}-S on three manipulation tasks: Place Pens, Store Block, and Stack Cups. We use approximately 1.1k demonstration trajectories in total and train for 63k optimization steps. Each method is evaluated over 20 trials per task, totaling 60 trials across the three tasks. As shown in Figure~\ref{fig:real_world} of the main paper, \modelname{}-S achieves success rates of 90\%, 85\%, and 90\%, respectively, compared with 80\%, 70\%, and 70\% for FastWAM. These rates correspond to 18/20, 17/20, and 18/20 successful trials for \modelname{}-S, compared with 16/20, 14/20, and 14/20 for FastWAM. The average success rate increases from 73.3\% (44/60) to 88.3\% (53/60), an absolute gain of 15.0 percentage points. The improvements across all three tasks support the effectiveness of joint state and visual difference modeling in this real-world setting. Broader evaluation across objects, environments, and embodiments remains necessary to establish generality.

\subsection{Visualization Results}
\label{app:visualization}

\subsubsection{Task Level Output Visualization}
\begin{figure*}[t]
    \centering
    \newcommand{\taskoutputgap}{6pt}
    \includegraphics[width=0.675\textwidth]{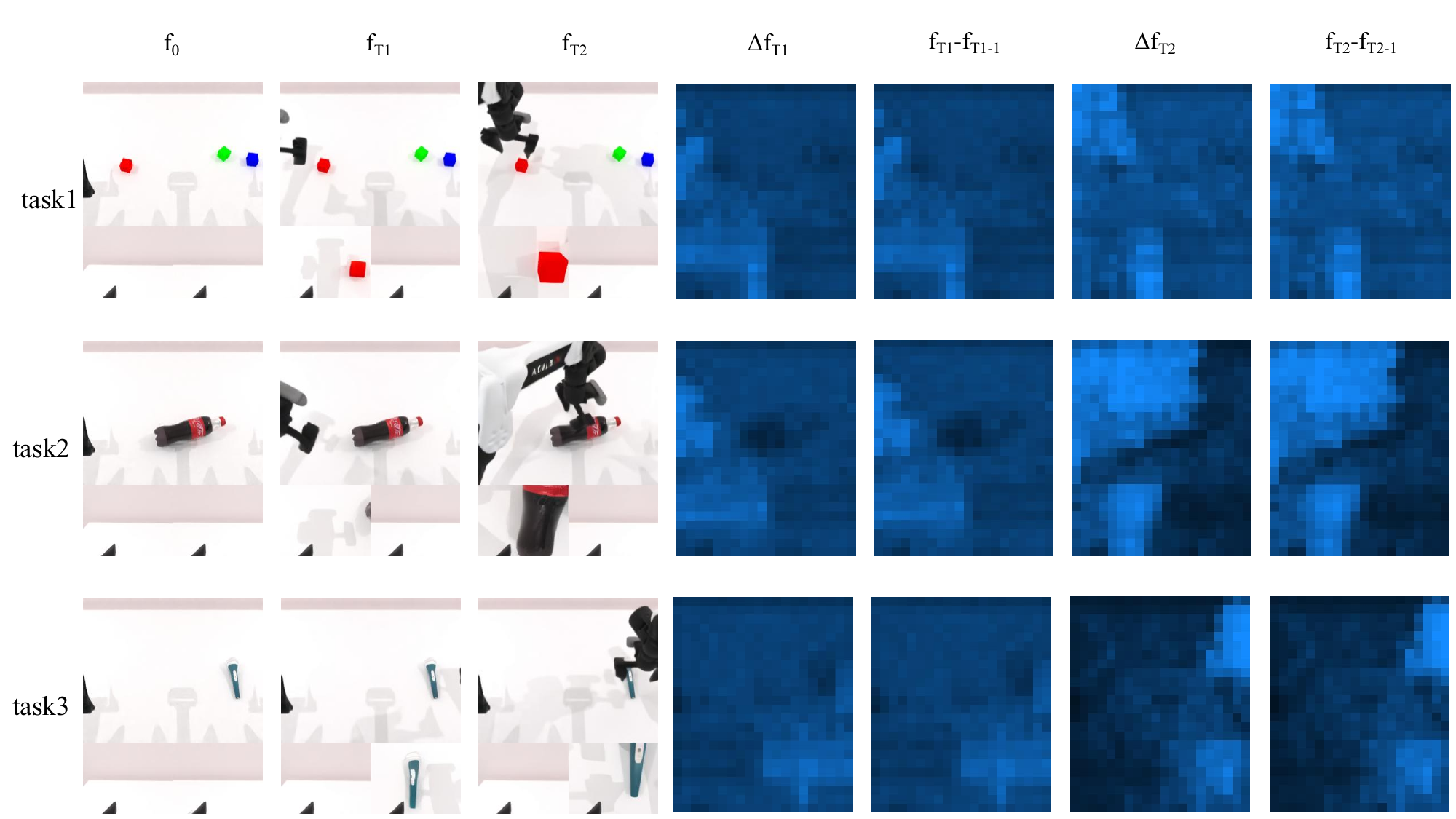}

    \vspace{\taskoutputgap}
    \includegraphics[width=0.675\textwidth]{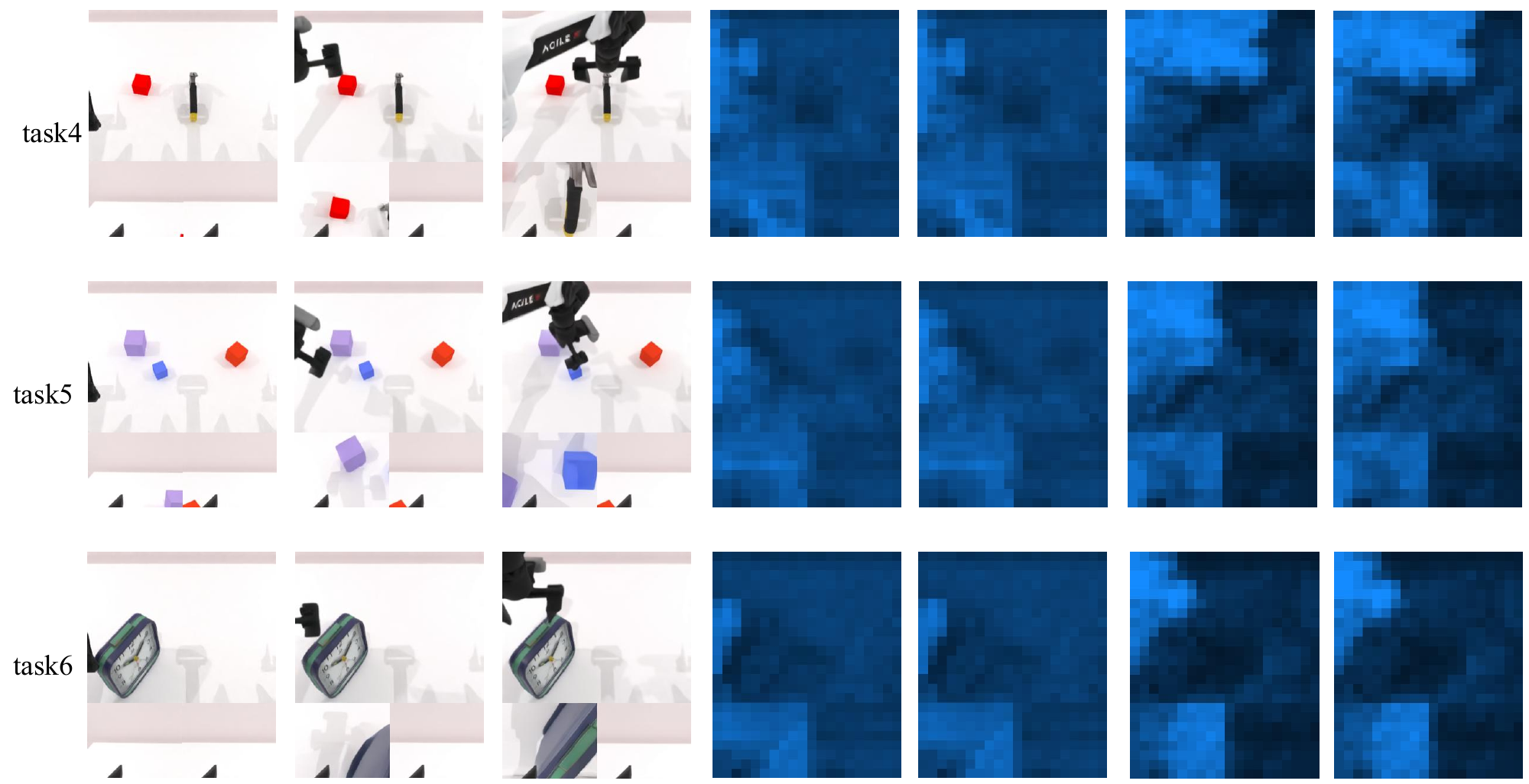}

    \vspace{\taskoutputgap}
    \includegraphics[width=0.675\textwidth]{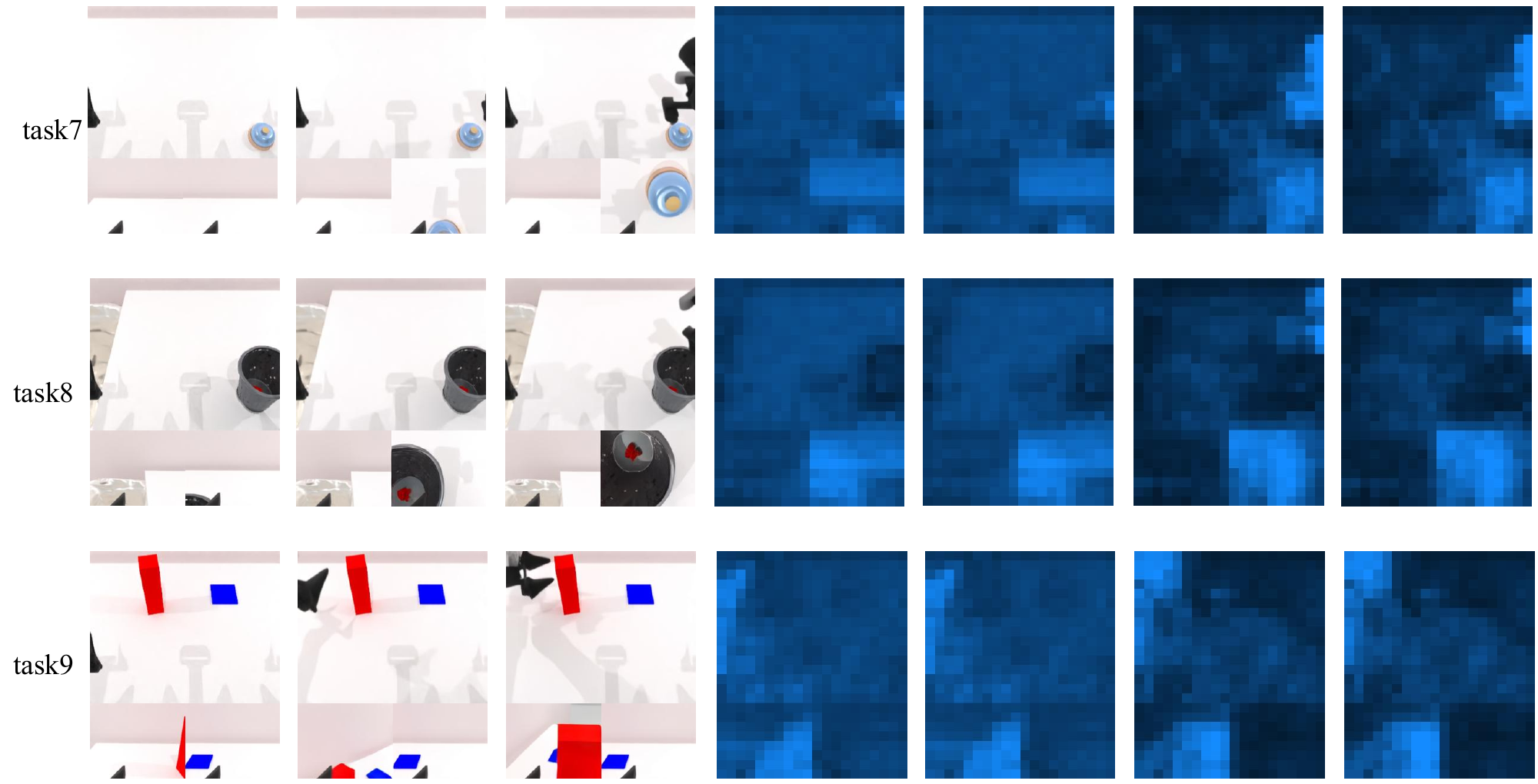}
    \caption{Qualitative visualization of future-state and visual difference prediction by the World and Diff Experts on representative manipulation tasks.}
    \label{fig:task_output}
\end{figure*}

Figure~\ref{fig:task_output} qualitatively visualizes the outputs of the World and Diff Experts on representative manipulation tasks. Both experts receive the same noised future-frame latents: the World Expert predicts future states through denoising, while the Diff Expert directly regresses differences between consecutive ground-truth frame latents. The predicted future observations preserve scene structure and reflect the progression of the manipulation trajectory. The visualized diff outputs highlight localized variation around the robot and manipulated objects, illustrating the complementary information captured by state and visual difference prediction. These differences describe inter-frame visual variation and do not exclusively isolate changes caused by robot actions.

\begin{table}[t]
    \centering
    {\small
    \setlength{\tabcolsep}{6pt}
    \renewcommand{\arraystretch}{1.12}
    \begin{adjustbox}{max width=\linewidth}
\begin{tabular}{lcccc}
        \toprule
        Metric & $f_1$ & $f_2$ & $\Delta f_1$ & $\Delta f_2$ \\
        \midrule
        PSNR$\Uparrow$ & 37.94 & 30.79 & - & - \\
        SSIM$\Uparrow$ & 0.9851 & 0.9701 & - & - \\
        Latent Error$\Downarrow$ & - & - & 0.0055 & 0.0136 \\
        \bottomrule
    \end{tabular}
\end{adjustbox}
    }
    \caption{Future-state and visual difference prediction. PSNR and SSIM evaluate predicted future observations; latent error evaluates predicted differences between consecutive frame latents.}
    \label{tab:future_delta}
\end{table}

Table~\ref{tab:future_delta} quantitatively evaluates the two prediction branches. For the first and second future observations, the World Expert achieves PSNR values of 37.94 and 30.79 and SSIM values of 0.9851 and 0.9701, respectively. The Diff Expert obtains latent prediction errors of 0.0055 and 0.0136 for $\Delta f_1$ and $\Delta f_2$. These results characterize the model's ability to predict both future scene states and their inter-frame latent~differences.

\subsection{Training Performance Analysis}

\begin{figure}[!htbp]
    \centering
    \includegraphics[width=\columnwidth]{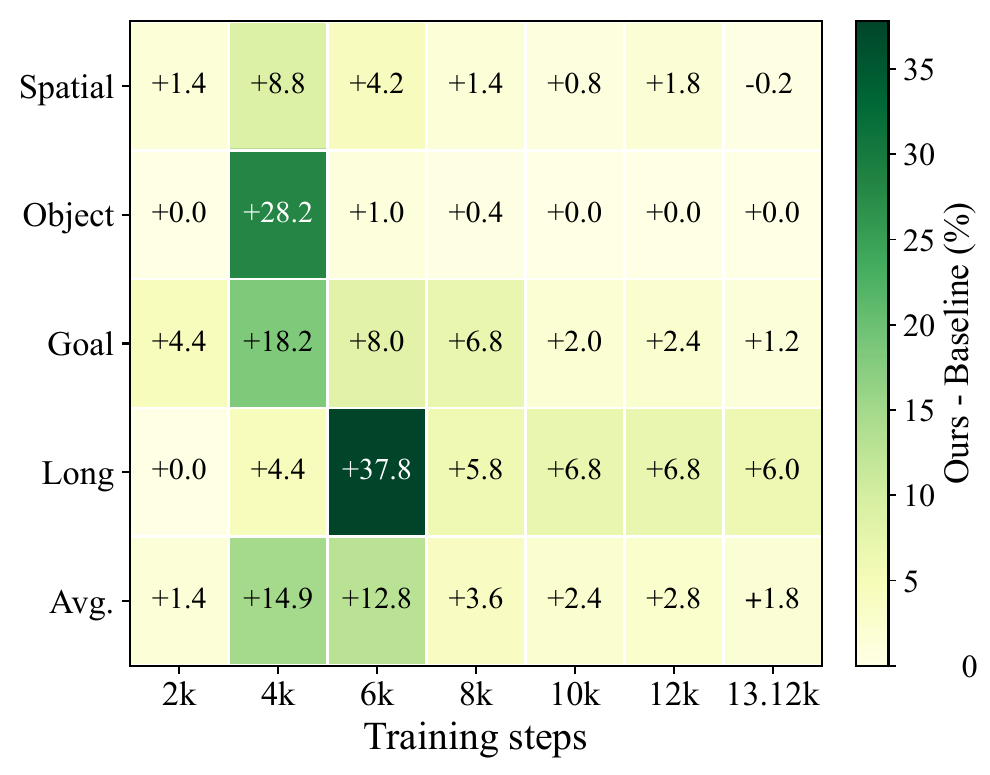}
    \caption{Supplementary per-suite training improvements of \modelname{} over the FastWAM baseline on LIBERO at each evaluation checkpoint.}
    \label{fig:training_performance_appendix}
\end{figure}

Figure~\ref{fig:training_performance_appendix} reports the per-suite improvement of \modelname{} over the FastWAM baseline throughout training. \modelname{} achieves non-negative gains at all earlier evaluation checkpoints and improves the final average success rate. The advantage is strongest in the early and middle stages: the average gain reaches 14.9 points at 4k steps and 12.8 points at 6k steps. At 4k steps, the improvements on LIBERO-Object and LIBERO-Goal are 28.2 and 18.2 points, respectively; LIBERO-Long reaches its largest gain of 37.8 points at 6k steps. Although the gaps narrow as training proceeds, \modelname{} improves the final average success rate by 1.75 percentage points, with gains on Goal and Long, a tie on Object, and a 0.2-point decrease on Spatial, including a 6.0-point gain on LIBERO-Long. These results show that \modelname{} provides consistent training improvements, especially for the more challenging goal-conditioned and long-horizon tasks.

\FloatBarrier
\clearpage
\nobalance
\section{Limitations and Future Work}
\label{app:limitations}

\subsection{Limitations}

Several limitations remain. First, the independent $M$-layer Diff Expert pathway introduces additional computation and memory during training. Freezing the LDB in \modelname{}-S does not eliminate this cost, since diff predictions are still computed and their gradients propagate into the shared representation. Both the World and Diff Experts are removed at inference, so these visual prediction pathways add no online action-generation overhead. Second, the difference target captures inter-frame latent variation without explicitly distinguishing changes caused by robot actions from camera motion, illumination changes, or movement of unrelated objects. This ambiguity may be more significant in dynamic, unstructured environments. Third, the clean-to-random result establishes a benefit under the evaluated RoboTwin perturbations, while absolute Random success remains low and generalization to other shifts is not established. The experiments cover a limited range of data scales and embodiments, and \modelname{} is not pretrained on a large-scale robot dataset. Broader evaluation is needed to establish how the benefits of \modelname{}-S and \modelname{}-L generalize across training-data regimes and deployment settings.

\newpage
\subsection{Future Work}

Future work will extend \modelname{} in three corresponding directions. First, more efficient Diff Expert designs, such as lightweight prediction heads or selective temporal token processing, could reduce the additional training cost and enable scaling to larger backbones, longer horizons, and richer multi-view observations. Second, we will investigate action-conditioned, motion-aware, and object-centric constraints to disentangle action-controllable changes from nuisance dynamics; multi-scale and multi-horizon diff prediction may further capture both fine-grained contact dynamics and longer-term task progress. Third, large-scale pretraining on diverse robot interaction data is a promising direction for learning more transferable visual and action priors before downstream adaptation. Evaluating this setting across different embodiments, camera configurations, objects, and long-horizon real-world tasks will be important for establishing the generality of diff prediction.

\FloatBarrier
\clearpage
\nobalance

\begin{table*}[t]
    \centering
    {\footnotesize
    \setlength{\tabcolsep}{1.3pt}
    \renewcommand{\arraystretch}{1.02}
    \begin{adjustbox}{max width=\linewidth}
\begin{tabular}{lcccccccccccccc}
        \toprule
        \multirow{2}{*}{Task} & \multicolumn{2}{c}{$\pi$0.5} & \multicolumn{2}{c}{Motus} & \multicolumn{2}{c}{LingBot-VA} & \multicolumn{2}{c}{GigaWorld-Policy} & \multicolumn{2}{c}{X-WAM} & \multicolumn{2}{c}{FastWAM} & \multicolumn{2}{c}{\modelname{}-L} \\
        \cmidrule(lr){2-3}\cmidrule(lr){4-5}\cmidrule(lr){6-7}\cmidrule(lr){8-9}\cmidrule(lr){10-11}\cmidrule(lr){12-13}\cmidrule(lr){14-15}
         & Clean & Random & Clean & Random & Clean & Random & Clean & Random & Clean & Random & Clean & Random & Clean & Random \\
        \midrule

        Adjust Bottle & \textbf{100} & \underline{99} & 89 & 93 & \underline{90} & 94 & \textbf{100} & \textbf{100} & \textbf{100} & \underline{99} & \textbf{100} & \textbf{100} & \textbf{100} & \textbf{100} \\
        Beat Block Hammer & 96 & 93 & 95 & 88 & 96 & \underline{98} & 86 & 86 & \underline{98} & 96 & \textbf{99} & 90 & 97 & \textbf{100} \\
        Blocks Ranking RGB & 92 & 85 & \underline{99} & 97 & \underline{99} & \underline{98} & 92 & 96 & \underline{99} & 95 & \textbf{100} & \textbf{100} & \textbf{100} & \textbf{100} \\
        Blocks Ranking Size & 49 & 26 & 75 & 63 & \underline{94} & 96 & 44 & 48 & 76 & 82 & \underline{94} & \textbf{100} & \textbf{100} & \underline{97} \\
        Click Alarmclock & 98 & 89 & \textbf{100} & \textbf{100} & \underline{99} & \textbf{100} & \textbf{100} & \textbf{100} & 98 & \underline{99} & \textbf{100} & \textbf{100} & \textbf{100} & \textbf{100} \\
        Click Bell & \underline{99} & \underline{66} & \textbf{100} & \textbf{100} & \textbf{100} & \textbf{100} & \textbf{100} & \textbf{100} & \textbf{100} & \textbf{100} & \textbf{100} & \textbf{100} & \textbf{100} & \textbf{100} \\
        Dump Bin Bigbin & 92 & \underline{97} & 95 & 91 & 89 & 96 & 92 & \textbf{100} & 90 & 96 & \underline{97} & \underline{97} & \textbf{100} & \textbf{100} \\
        Grab Roller & \textbf{100} & \textbf{100} & \textbf{100} & \textbf{100} & \textbf{100} & \textbf{100} & \textbf{100} & \textbf{100} & \textbf{100} & \textbf{100} & \textbf{100} & \textbf{100} & \textbf{100} & \textbf{100} \\
        Handover Block & 66 & 57 & 86 & 73 & \textbf{99} & 78 & 80 & 80 & 88 & 79 & 95 & \underline{83} & \underline{97} & \textbf{90} \\
        Handover Mic & 98 & \underline{97} & 78 & 63 & 94 & 96 & 72 & 72 & 88 & 89 & \underline{99} & \textbf{100} & \textbf{100} & \textbf{100} \\
        Lift Pot & 18 & 17 & 38 & 38 & 40 & 28 & 16 & 12 & \textbf{100} & \underline{99} & \underline{58} & 97 & \textbf{100} & \textbf{100} \\
        Move Can Pot & 96 & 85 & 96 & \textbf{99} & \textbf{100} & \textbf{99} & \underline{98} & \underline{98} & 80 & 84 & \textbf{100} & 80 & 90 & 87 \\
        Move Playingcard Away & 51 & 55 & 34 & 74 & \underline{94} & 97 & 76 & 78 & \textbf{100} & \underline{98} & 90 & \textbf{100} & \textbf{100} & \textbf{100} \\
        Move Stapler Pad & 84 & 61 & 93 & \underline{96} & \underline{99} & \textbf{99} & 90 & 90 & 67 & 70 & \textbf{100} & 53 & 73 & 57 \\
        Hanging Mug & \underline{96} & 84 & \textbf{100} & \underline{96} & \textbf{100} & \textbf{99} & 78 & 72 & 46 & 55 & \textbf{100} & 53 & 63 & 57 \\
        Open Laptop & 56 & 42 & 83 & 85 & 91 & 79 & 92 & 82 & \underline{96} & \underline{97} & 77 & \textbf{100} & \textbf{100} & \textbf{100} \\
        Open Microwave & 90 & \underline{96} & 95 & 91 & 92 & 94 & \underline{96} & \textbf{98} & 89 & 92 & \textbf{98} & 37 & 73 & 53 \\
        Pick Diverse Bottles & 34 & 77 & \textbf{95} & \underline{91} & 82 & 86 & 74 & 66 & 91 & \textbf{92} & 62 & 87 & \underline{93} & 87 \\
        Pick Dual Bottles & 81 & 71 & 90 & 91 & 89 & 82 & 82 & 70 & \underline{99} & \textbf{100} & 80 & \underline{97} & \textbf{100} & \underline{97} \\
        Place A2B Left & 93 & 63 & \underline{96} & 90 & \textbf{100} & \textbf{99} & 86 & 86 & 90 & 87 & \textbf{100} & 93 & 93 & \underline{97} \\
        Place A2B Right & 87 & 82 & 88 & 79 & \textbf{97} & 93 & 94 & 88 & 92 & 89 & \underline{95} & \underline{97} & 90 & \textbf{100} \\
        Place Bread Basket & 87 & 84 & 91 & 87 & \textbf{97} & \textbf{95} & 90 & \underline{92} & 90 & 91 & \underline{93} & 87 & 87 & 90 \\
        Place Bread Skillet & 77 & 64 & \underline{91} & 94 & \textbf{97} & 95 & 82 & 82 & 90 & \underline{96} & \underline{91} & \textbf{97} & \textbf{97} & 93 \\
        Place Can Basket & 85 & 66 & 86 & \underline{83} & \textbf{95} & \textbf{90} & \underline{94} & \textbf{90} & 84 & 82 & 90 & 60 & 63 & 77 \\
        Place Cans Plasticbox & 94 & 87 & 98 & \underline{98} & 97 & 95 & 98 & 96 & \underline{99} & \underline{98} & 71 & \textbf{100} & \textbf{100} & \textbf{100} \\
        Place Container Plate & 62 & 62 & 81 & 76 & 81 & 84 & 78 & 74 & 98 & \textbf{100} & \underline{99} & \underline{97} & \textbf{100} & \textbf{100} \\
        Place Dual Shoes & 94 & 84 & \underline{98} & 94 & \textbf{100} & \underline{99} & \textbf{100} & \textbf{100} & 83 & 81 & 96 & 87 & 93 & 87 \\
        Place Empty Cup & \underline{99} & 95 & 98 & \underline{99} & \underline{99} & 97 & 98 & 96 & 98 & \underline{99} & 94 & \textbf{100} & \textbf{100} & \textbf{100} \\
        Place Fan & 75 & 75 & 93 & 87 & 94 & 89 & 96 & 84 & 84 & 92 & \textbf{100} & \textbf{97} & \underline{97} & \underline{93} \\
        Place Burger Fries & \textbf{100} & \underline{99} & \underline{99} & 98 & \textbf{100} & \textbf{100} & 90 & 90 & 97 & \underline{99} & 96 & \textbf{100} & \textbf{100} & 97 \\
        Place Mouse Pad & 87 & 85 & 91 & 87 & \textbf{99} & \underline{93} & 92 & \textbf{94} & 84 & 86 & \underline{96} & 87 & 87 & 83 \\
        Place Object Basket & 60 & 39 & 66 & 68 & \underline{93} & \textbf{96} & 88 & \underline{90} & 85 & 87 & 83 & 77 & \textbf{97} & 87 \\
        Place Object Scale & 80 & 76 & 81 & 87 & \underline{91} & 88 & 90 & 92 & \textbf{93} & 89 & 89 & \underline{93} & \textbf{93} & \textbf{97} \\
        Place Object Stand & 86 & 80 & 88 & 85 & \underline{96} & \underline{95} & 88 & 80 & \textbf{97} & \textbf{96} & 90 & 90 & 93 & 87 \\
        Place Phone Stand & 91 & 85 & 98 & \underline{97} & \underline{99} & 96 & \textbf{100} & \textbf{98} & 75 & 80 & 90 & \underline{97} & 97 & \underline{97} \\
        Move Pillbottle Pad & 81 & 81 & 87 & 86 & \textbf{97} & \underline{97} & 82 & 72 & \underline{96} & \textbf{98} & \textbf{97} & 90 & \textbf{97} & \underline{97} \\
        Place Shoe & 92 & 93 & \textbf{99} & 97 & \underline{98} & \underline{98} & \underline{98} & 96 & 97 & \textbf{99} & 96 & 97 & 97 & 97 \\
        Press Stapler & 87 & 83 & 93 & \underline{98} & 85 & 82 & \underline{96} & 96 & 94 & 90 & 90 & \textbf{100} & \textbf{97} & \textbf{100} \\
        Put Bottles Dustbin & 84 & 79 & 81 & 79 & \underline{87} & \underline{91} & 72 & 70 & 85 & \textbf{95} & \textbf{95} & 83 & \underline{87} & 83 \\
        Put Object Cabinet & 80 & 79 & 88 & 71 & 85 & 87 & 74 & 74 & 66 & 76 & \textbf{94} & \textbf{93} & \underline{93} & \underline{90} \\
        Rotate Qrcode & 89 & 87 & 89 & 73 & \textbf{96} & \textbf{91} & 90 & 84 & 84 & 83 & \underline{93} & \underline{90} & 87 & \underline{90} \\
        Scan Object & 72 & 65 & 67 & 66 & \underline{96} & \underline{91} & 60 & 64 & 86 & 79 & 89 & \textbf{93} & \textbf{97} & 87 \\
        Shake Bottle & \underline{99} & 97 & \textbf{100} & 97 & \textbf{100} & 97 & \textbf{100} & 98 & \underline{99} & \underline{99} & \textbf{100} & \textbf{100} & \textbf{100} & \textbf{100} \\
        Shake Bottle Horizontally & \underline{99} & \underline{99} & \textbf{100} & 98 & \textbf{100} & \underline{99} & \textbf{100} & \textbf{100} & \textbf{100} & \underline{99} & \textbf{100} & \textbf{100} & \textbf{100} & \textbf{100} \\
        Stack Blocks Three & 91 & 76 & 91 & 95 & \textbf{99} & \underline{98} & 70 & 78 & \underline{97} & 95 & 95 & 97 & 93 & \textbf{100} \\
        Stack Blocks Two & \underline{97} & \textbf{100} & \textbf{100} & \underline{98} & \textbf{100} & \underline{98} & \textbf{100} & 94 & \textbf{100} & \textbf{100} & \textbf{100} & \textbf{100} & \textbf{100} & \textbf{100} \\
        Stack Bowls Three & 77 & 71 & 79 & \textbf{87} & \underline{86} & \underline{83} & 70 & 72 & \textbf{88} & 82 & 80 & \textbf{87} & 77 & \textbf{87} \\
        Stack Bowls Two & 95 & 96 & \textbf{98} & \underline{98} & 94 & \underline{98} & \underline{96} & 92 & \textbf{98} & \underline{98} & 92 & \textbf{100} & 93 & 97 \\
        Stamp Seal & 79 & 55 & \underline{93} & 92 & \textbf{96} & \underline{97} & \textbf{96} & \textbf{98} & \underline{93} & 95 & 90 & 93 & \underline{93} & 93 \\
        Turn Switch & 62 & 54 & \textbf{84} & \underline{78} & 44 & 45 & \underline{82} & \textbf{84} & 61 & 72 & 61 & 60 & 80 & 70 \\
        \midrule
        \textbf{Overall} & 82.74 & 76.76 & 88.66 & 87.02 & \underline{92.90} & \underline{91.50} & 86.36 & 85.04 & 89.76 & 90.68 & 91.88 & 90.32 & \textbf{93.28} & \textbf{92.02} \\
        \bottomrule
    \end{tabular}
\end{adjustbox}
    }
    \caption{Task-level RoboTwin 2.0 results. Columns report clean/random success rates for each method.}
    \label{tab:robotwin_task_full}
\end{table*}

\FloatBarrier

\end{document}